\documentclass[sigconf]{acmart}
\usepackage{amsfonts}
\usepackage{booktabs} % For formal tables
\usepackage{subfigure}
\usepackage{bm}
\usepackage{xspace}
\usepackage{multirow}
\usepackage{algorithm}
\usepackage{algpseudocode}
\usepackage[normalem]{ulem}
\usepackage{array}
\usepackage{graphicx}
\usepackage{multirow}
\usepackage{enumitem}

\usepackage{soul}
\usepackage{epstopdf}
\usepackage{setspace}
\usepackage{pifont}
\usepackage{xcolor}
\usepackage{mathrsfs}
\usepackage{amsmath}
\usepackage{ulem}
\usepackage{float}

\newcommand{\paratitle}[1]{\vspace{1.5ex}\noindent\textbf{#1}}
\newcommand{\ie}{\emph{i.e.,}\xspace}

\newcommand{\eg}{\emph{e.g.,}\xspace}

\newcommand{\ignore}[1]{}

\newcommand{\tabincell}[2]{\begin{tabular}{@{}#1@{}}#2\end{tabular}}

\AtBeginDocument{%
  \providecommand\BibTeX{{%
    \normalfont B\kern-0.5em{\scshape i\kern-0.25em b}\kern-0.8em\TeX}}}

\copyrightyear{2021}
\acmYear{2021}
\setcopyright{acmcopyright}\acmConference[SIGIR '21]{Proceedings of the 44th International ACM SIGIR Conference on Research and Development in Information Retrieval}{July 11--15, 2021}{Virtual Event, Canada}
\acmBooktitle{Proceedings of the 44th International ACM SIGIR Conference on Research and Development in Information Retrieval (SIGIR '21), July 11--15, 2021, Virtual Event, Canada}
\acmPrice{15.00}
\acmDOI{10.1145/3404835.3462865}
\acmISBN{978-1-4503-8037-9/21/07}

\begin{document}
\fancyhead{}
\title{Knowledge-based Review Generation by \\Coherence Enhanced Text Planning}

\author{Junyi Li$^{1,2}$, Wayne Xin Zhao$^{1,2*}$, Zhicheng Wei$^{3}$, Nicholas Jing Yuan$^{3}$ and Ji-Rong Wen$^{1,2}$}
\thanks{$^*$Corresponding author.}
\affiliation{%
	\institution{$^1$Gaoling School of Artificial Intelligence, Renmin University of China}
	\institution{$^2$Beijing Key Laboratory of Big Data Management and Analysis Methods}
	\institution{$^3$Huawei Cloud}
}
\affiliation{%
	\institution{\{lijunyi,jrwen\}@ruc.edu.cn, \{batmanfly, nicholas.jing.yuan\}@gmail.com, weizhicheng1@huawei.com}
}

\begin{abstract}
As a natural language generation task, it is challenging to generate informative and coherent review text. In order to enhance
the informativeness of the generated text, existing solutions typically learn to copy entities or triples from knowledge graphs (KGs).
However, they lack overall consideration to select and arrange the
incorporated knowledge, which tends to cause text incoherence.

To address the above issue, we focus on improving entity-centric
coherence of the generated reviews by leveraging the semantic structure of KGs. In this paper, we propose a novel Coherence Enhanced
Text Planning model (CETP) based on knowledge graphs (KGs) to
improve both global and local coherence for review generation. The
proposed model learns a two-level text plan for generating a document: (1) the document plan is modeled as a sequence of sentence
plans in order, and (2) the sentence plan is modeled as an entity-based subgraph from KG. Local coherence can be naturally enforced
by KG subgraphs through intra-sentence correlations between entities. For global coherence, we design a hierarchical self-attentive
architecture with both subgraph- and node-level attention to enhance the correlations between subgraphs. To our knowledge, we
are the first to utilize a KG-based text planning model to enhance
text coherence for review generation. Extensive experiments on
three datasets confirm the effectiveness of our model on improving
the content coherence of generated texts.
\end{abstract}




\maketitle

\section{Introduction}
With the development of e-commerce in recent years, online reviews play a crucial role in reflecting real customer experiences. Online review information is useful to both users interested in a certain product and sellers concerned about increasing their revenue. However, many users find it tedious to write review text, and a large proportion of users do not post online reviews~\cite{bareket2016strategic}. To ease the process of review writing, the task of review generation has been proposed and received wide attention from both research and industry communities~\cite{ZhouLWDHX17,NiM18,LiZWS19}. Review generation aims to automatically produce review text conditioned on some necessary context inputs (\eg users, items and ratings), which potentially influences many applications, such as explanation generation for recommendation~\cite{LiZC20}, automatic scientific reviewing for papers~\cite{BartoliLMT16}. Existing methods mainly  make  extensions based on sequential neural networks (\eg recurrent neural network), including attribute awareness~\cite{ZhouLWDHX17}, aspect enrichment~\cite{NiM18}, and length enhancement~\cite{LiZWS19}. While, these studies do not explicitly utilize the factual information about items, tending to generate dull and uninformative review text. 

\begin{figure}[t]
	\centering
	\includegraphics[width=0.45\textwidth]{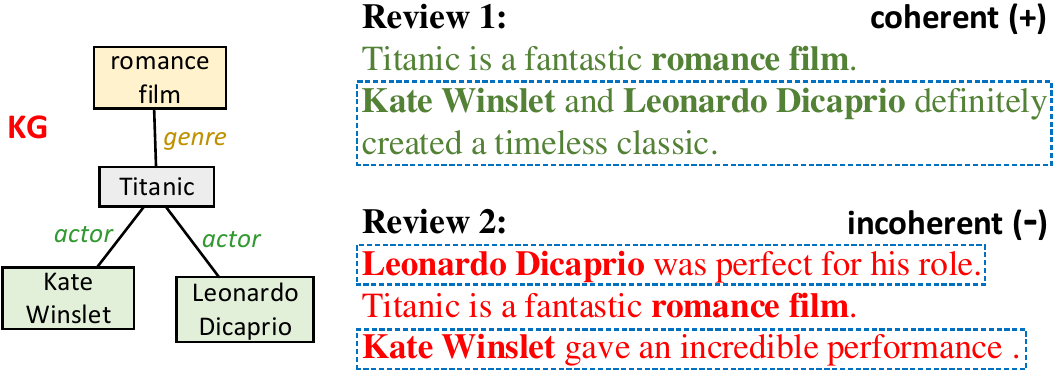}
	\caption{Comparison of coherent and incoherent review texts in terms of content continuity. Review 1 first comments on the movie genre, followed a sentence about two  actors; Review 2 introduces the two actors with two separate sentences, interleaved by another sentence on genre.}
	\label{fig-example}
	%\vspace{-0.3cm}
\end{figure}

To enrich the generated content, we consider incorporating external knowledge graph (KG) to improve review generation. % following~\cite{li2020knowledge}. 
By associating KG entities with e-commerce items, we can obtain rich semantic relations about items from KG. Indeed, there has also been growing interest in the utilization of KG data in other text generation tasks, such as dialog system~\cite{niu2019knowledge} and document summarization~\cite{AmplayoLH18}. 
These approaches typically learn to copy entities or triples from KG when necessary, which improves the informativeness of the generated content to a certain extent. However, they lack overall consideration to select and arrange the incorporated KG data, which is likely to cause the issue of text incoherence~\cite{GroszJW95}, such as content discontinuity and logic confusion. 
Figure~\ref{fig-example} presents a comparison between a coherent review and an incoherent review in terms of content continuity. As we can see, although review 2 has incorporated factual information from KG, the entire organization is poor and the review content is discontinuous in terms of semantic structure, \ie the lead actor and actress, \emph{Leonardo Dicaprio} and \emph{Kate Winslet}, are separated by the genre \emph{romance film}.
%As we can see, simply incorporating KG data will make  
%Intuitively, a user will only mention a small number of facts that she/he is interested in, and organized them into a coherent text. 
%Therefore, simply incorporating KG data is likely to cause content coherence issue~\cite{GroszJW95,KaramanisPMO04}, \eg semantic redundancy or conflicts. 

To address the above issue, we focus on improving \emph{entity-centric coherence} of the generated reviews by leveraging the semantic structure of KGs.
According to \cite{GroszJW95}, entity-centric coherence refers to the entities are closely correlated to the other in text and the entity correlations among sentences can be used to create coherence patterns for text.
Also, it has been widely recognized that entity graphs (or subgraphs) are a powerful form to characterize the coherent structure in natural language~\cite{GuinaudeauS13}. 
Compared with the traditional entity-grid method~\cite{BarzilayL05} which is restricted to capturing coherent transitions between adjacent sentences, the KG-based entity graphs can easily span the entire text and capture the semantic correlations between different sentences.
%\textcolor{blue}{Empirically, graphs are preferred for modeling coherence patterns due to their long-distance connectivity characteristics~\cite{GuinaudeauS13}.}
%Based on this motivation, we propose to utilize the semantic structure of KGs for enhancing text coherence of the generated reviews.
Following the literature of linguistics~\cite{mani1998using}, we consider two kinds of coherence, namely \emph{global} and \emph{local coherence}.  \emph{Global coherence} captures how entities distribute among different sentences through inter-sentence correlations~\cite{ElsnerAC07}. 
While \emph{local coherence} means the intra-sentence entities (or keywords) have close correlations through semantic relations or threading words~\cite{BarzilayL05}.
Our main idea  is aimed at utilizing  KG subgraphs and their correlations to enhance local and global coherence, respectively. 
To develop our model, we derive the generation plan before generating the text. 
Such a way is called \emph{text planning} in text generation~\cite{MoryossefGD19,HuaW19}, which refers to the process of selecting, arranging and ordering content to be produced.

To this end, in this paper, we propose  a novel \textbf{C}oherence \textbf{E}nhanced \textbf{T}ext \textbf{P}lanning model (CETP) for review generation. 
We utilize  KGs for capturing coherence patterns to generate coherent review text. 
%Specially, we extend original KGs (called \emph{HKG}) by incorporating user nodes and keywords for capturing personalized preference and entity-word associations. 
Based on an augmented KG with user-item interaction and entity-keyword co-occurrence, we design a two-level text plan for generating a document: (1) the  document plan
is modeled as a sequence of sentence plans in order, and (2) the sentence plan is modeled as an entity-based subgraph  from KG.
%, specifying the relation and information to be realized.
In our model, local coherence is naturally enforced by KG subgraphs, since entities from a KG subgraph are tightly associated by semantic relations. To enhance global coherence, we develop a hierarchical self-attentive architecture with both subgraph- and node-level attention to generate a coherent sequence of subgraphs. 
For sentence realization, we  develop a supervised copy mechanism
for copying entities (or keywords) from the planned subgraph. It  further improves local coherence by enhancing  the intra-sentence entity correlation via threading words. %In this way, our model can generate globally and locally coherent review text by utilizing the KG information.

%Therefore, to enhance global coherence, we develop a Transformer decoder with two-level attention to generate subgraph sequence from HKG. Each subgraph contains the entities and words to be generated in each sentence, which can capture the distribution of entities among sentences. In the word-level planning, a supervised copy mechanism is performed on Transformer decoder to explicitly guide the generation of entities and related words by coping them from subgraphs, which can improve the intra-sentence connectivity across entities. In this way, our model can generate globally and locally coherent review text by utilizing the KG information.

To the best of our knowledge, we are the first to utilize a KG-based text planning model to enhance both global and local coherence for review generation.
For evaluation, we construct three review datasets by associating KG entities with e-commerce items to obtain semantic attributes about items. Extensive experiments demonstrate the effectiveness of our model.

\section{Related Work}

With the striking success of deep neural networks, automatic review generation has received much attention from the research community~\cite{ZhouLWDHX17,NiM18,LiZWS19}.
Typical methods extend the Sequence-to-Sequence framework~\cite{SutskeverVL14} by using available side information, including context information~\cite{ZhouLWDHX17}, sentiment score~\cite{ZangW17} and user-item interactions~\cite{NiLVM17}. 
In order to alleviate the repetitiveness of texts caused by the RNN models, Generative Adversarial Nets (GAN) based approaches have been applied to text generation~\cite{YuZWY17,GuoLCZYW18}. 
Moreover, several studies utilize aspect information of products or writing style of users with a more instructive generation process to generate personalized or controllable review text~\cite{NiM18,LiZWS19,li2020knowledge}.
Although various  approaches have emerged, they seldom include structured attribute information about items, thus making the generated review text less informative. 

In many applications~\cite{AmplayoLH18,li2020knowledge,niu2019knowledge}, various approaches utilize structural knowledge data (\eg Freebase and DBpedia) in the text generation process in order to improve the informativeness and diversity of the generated content. However, most of these studies mainly learn to copy related entities or triples from structural knowledge data while lack overall consideration of semantic structure of text, which have limitation in generating semantically coherent text. 

Coherence is a key property of well-organized text. A variety of coherence analysis methods have been developed, such as entity grid model~\cite{BarzilayL05}, coherence pattern model~\cite{ParveenM016}, and neural network model~\cite{LiH14a}. However, these methods still require considerable experience or domain expertise to define or extract features. Other related approaches include global graph model~\cite{GuinaudeauS13} which projects entities into a global graph and HMM system~\cite{LouisN12} in which the coherence between adjacent sentences is modeled by a hidden Markov framework and captured by the transition rules of  topics.
%and pay little attention to essential semantic correlations among entities.

Text planning is a critical component in traditional data-to-text systems~\cite{MoryossefGD19,HuaW19}. Typical methods are based on hand-crafted~\cite{DalianisH93} or automatically-learnt rules~\cite{DuboueM03}, which are unable to capture rich variations of texts. Recent neural approaches mainly rely on well-designed network architectures  to learn from training data~\cite{ShaMLPLCS18,HuaW19}, which is difficult to control the planning process. However, as demonstrated in~\cite{WisemanSR17}, existing neural methods are still problematic for text generation and often generate incoherent text. Moreover, these text planning methods seldom focus on review generation task, lacking consideration of user-item interaction or personalized characteristics. 

\section{Problem Formulation}
\label{sec-preliminary}

In this section, we introduce the notations that will be used throughout the paper, and then formally define the task. 

\paratitle{Basic Notations}.
Let $\mathcal{U}$, $\mathcal{I}$ and $\mathcal{A}$ denote a user set, an item set and a rating score set, respectively. 
A review text is written by a user $u \in \mathcal{U}$ about an item $i \in \mathcal{I}$ with a rating score of $a \in \mathcal{A}$. 
Formally, a review text  is denoted by $w^{1:m}=\{\langle w_{j,1},\cdots,w_{j,t},\cdots,w_{j,n_j} \rangle \}_{j=1}^m$, consisting of $m$ sentences, where $w_{j,t}$ denotes the $t$-th word (from the vocabulary $\mathcal{V}$) of the $j$-th sentence and $n_j$ is the length of the $j$-th sentence. Besides, in our setting, a knowledge graph~(KG) $\mathcal{T}$ about item attributes is available for our task. Typically, it organizes facts by triples:  $\mathcal{T}=\{\langle h,r,t \rangle \}$, where each triple describes that there is a relation $r$ between head entity $h$ and tail entity $t$ regarding to some facts. We assume that a KG entity  can be aligned to an e-commerce item.
For instance, the Freebase movie entity  ``\emph{Avatar}'' (with the Freebase ID \emph{m.0bth54}) has an entry of a movie item in IMDb (with the IMDb ID \emph{tt0499549}). Several studies~\cite{zhao2019kb4rec,li2020knowledge} try to develop heuristic algorithms for  item-to-entity alignment and have released public linkage dataset.
%Following~\cite{li2020knowledge}, 
We add user-item links according to their interactions and entity-keyword links if they frequently co-occur in review sentences, in order to capture personalized entity preference and enhance the entity-word associations, respectively. 
For unifying the triple form, two kinds of non-KG links are attached with two new relations, \ie interaction and co-occurrence.
%For the two kinds of additional links, we further incorporate two new relations, \ie interaction and description, to unify the triple form.
As shown in Figure~\ref{fig-model}(c), such an augmented KG can be referred as \emph{Heterogeneous KG}~(HKG), denoted by $\mathcal{G} = \mathcal{T} \cup \{\langle u, r_{int}, i \rangle\} \cup \{\langle e, r_{co}, w \rangle \}$, where $r_{int}$ and $r_{co}$ denote the relations of user-item interaction and entity-word co-occurrence, respectively.

\paratitle{Planning with HKG Subgraphs}. 
To enhance the global and local coherence, we design a two-level text plan for selecting, arranging and ordering the contents in the output text, namely \emph{document plan} and \emph{sentence plan}. Specifically, the  document plan is modeled as a sequence of sentence plans in order, denoted by $g^{1:m}={\langle g_1,\cdots,g_j,\cdots,g_m \rangle}$. Each sentence plan is modeled as an entity-based subgraph $g_j$ (short for \emph{subgraph}) from the HKG shown in Figure~\ref{fig-model}(b), specifying the relations and entities (or keywords) to be verbalized in each sentence. We further introduce the concept of \emph{subgraph schema} denoted by $s_j$ for subgraph $g_j$, as shown in Figure~\ref{fig-model}(a), which keeps the structure and relations of subgraphs while replaces nodes with empty slots. Typically, a subgraph schema can be instantiated into different subgraphs by filling empty slots with different entities or keywords.

\paratitle{Task Definition}. 
Review generation task is concerned with how to automatically generate the review text $w^{1:m}$ for a rating record $\langle u, i, a\rangle$ based on other possible side information if any. Different from most of previous works, we incorporate the HKG $\mathcal{G}$ as available resource for review generation.
We would like to utilize KG-based text planning model to enhance global and local coherence of the generated text.

\section{The Proposed Approach}
\begin{figure*}[t]
	\centering 
	\includegraphics[width=1\textwidth]{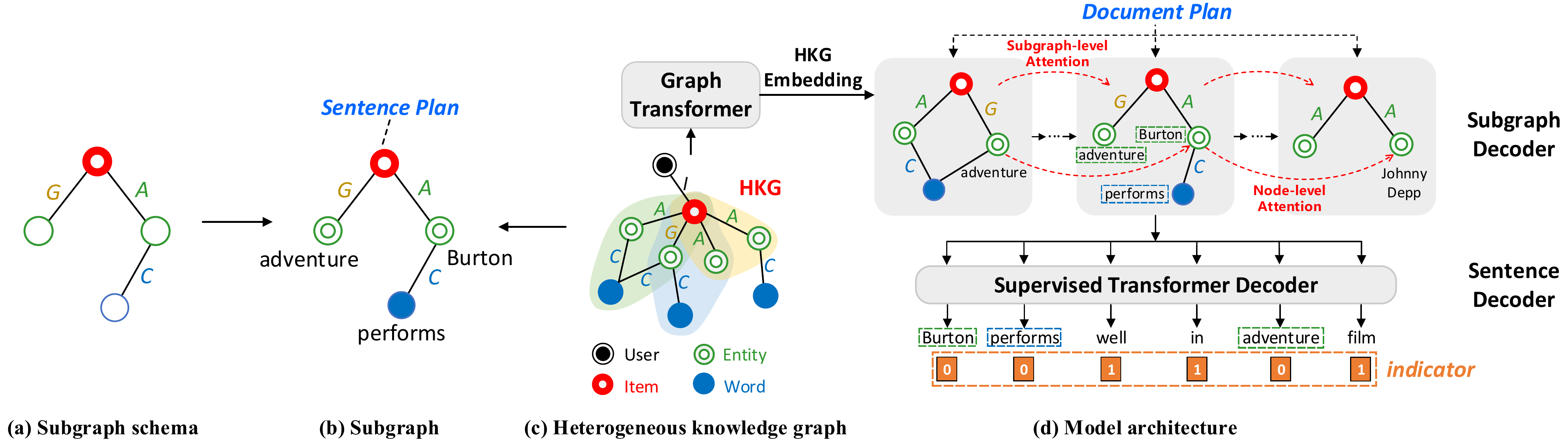} 
	\caption{Illustrative examples for our approach: (a) subgraph schema (the capital letters $I$, $A$, $G$ and $C$ denote the relations of \emph{interaction}, \emph{actor}, \emph{genre} and \emph{co-occurrence}, respectively); (b) subgraph; (c) heterogeneous knowledge graph; and (d) the overview of the proposed CETP model. The document plan and sentence plan can enhance the global and local coherence, respectively.} 
	\label{fig-model} 
\end{figure*}

In this section, we present the proposed \underline{C}oherence \underline{E}nhanced \underline{T}ext \underline{P}lanning model, named \emph{CETP}, for  the review generation task. We first introduce the two-level text plan for selecting, arranging and ordering the contents in the output text, namely \emph{document plan} and \emph{sentence plan}. 
As discussed earlier, a document plan is modeled as a sequence of sentence plans in order and a sentence plan is modeled as an entity-based subgraph from KG. Subgraphs naturally enforce the \emph{local coherence} of entities in a sentence, since they are originally connected and associated with relations in HKG. Furthermore, subgraph sequence can capture the overall  distribution and arrangement of entities, which helps improve \emph{global coherence}. 
Based on the two-level text plan, we adopt a supervised copy mechanism for sentence realization by copying entities (or keywords) from the planned subgraph. This step further improves local coherence by enhancing the intra-sentence entity correlation via threading words.
Figure~\ref{fig-model} presents examples for illustrating the basic notations and coherence enhancement in our model. 

\subsection{Text Plan Generation}

In this step, we study how to generate the text plan, \ie a subgraph sequence $g^{1:m}={\langle g_1,\cdots,g_j,\cdots,g_m \rangle}$ defined in Section~\ref{sec-preliminary}. 
%We extend the recently proposed 

\subsubsection{HKG Embedding} 
\label{sec-graph-encoding}

%In order to capture the semantic correlations between nodes, we propose to use Relation-Enhanced Graph Transformer~\cite{DengC20}, which considers the relation path between nodes. Our graph encoder stacks self-attention block by $L$ times, and each block consists of two sub-layers: relation-enhanced multi-head attention and feed-forward network.
We first learn node representations for the HKG. 
Let $n$ ($n_j$ and $n_k$) denote a node placeholder for the HKG, associated with an embedding vector $\bm{v}_n \in \mathbb{R}^{d_E}$, where $d_E$ denotes the node embedding size. Node embeddings can be initialized with pre-trained KG embeddings or word embeddings~\cite{YangYHGD14a,MikolovSCCD13}. In order to capture the semantic correlations between nodes,
we propose to use Relation-Enhanced Graph Transformer~\cite{VaswaniSPUJGKP17}, %which considers the relation path between nodes.
%The node embedding $\bm{v}_n \in \mathbb{R}^{d_E}$ for entity or word node $n$ can be initialized with pre-trained KG embeddings or word embeddings.
%In order to capture the semantic correlation between nodes, we propose to use Relation-Enhanced Graph Transformer~\cite{DengC20}, which considers the relation path between nodes. Our graph encoder stacks self-attention block by $L$ times, and each block consists of two sub-layers: relation-enhanced multi-head attention and feed-forward network.
%Firstly, we treat $\mathcal{G}$ as a fully connected graph and apply the relation-enhanced multi-head attention.
which applies a relation-enhanced multi-head attention (MHA) to obtain the node embedding $\bm{\hat{v}}_{n_j}$ for node $n_j$ as:
\begin{equation}\label{eq-encoder-multi-head}
\bm{\hat{v}}_{n_j} = \underset{n_k \in \mathcal{G}}{\text{MHA}}(\bm{q}_{n_j}, \bm{k}_{n_k}, \bm{v}_{n_k}),
\end{equation}
where $\text{MHA}(\bm{Q}, \bm{K}, \bm{V})$ follows the implementation of  multi-head attention~\cite{VaswaniSPUJGKP17} taking a query $\bm{Q}$, a key $\bm{K}$ and a value $\bm{V}$ as input: 
\begin{eqnarray}\label{eq-multi-head-define}
\text{MHA}(\bm{Q}, \bm{K}, \bm{V}) &=& \text{Concat}(\text{head}_1, ..., \text{head}_H)\bm{W}^O, \\
\text{head}_h &=& \text{Attn}(\bm{Q}\bm{W}^Q_h, \bm{K}\bm{W}^K_h, \bm{V}\bm{W}^V_h), \nonumber 
%\text{Attn}(\bm{Q}_h, \bm{K}_h, \bm{V}_h) &=& \text{softmax}(\frac{\bm{Q}_h\bm{K}_h^\top}{\sqrt{d_H}})\bm{V}_h, \nonumber 
\end{eqnarray}
%where $\bm{Q}_h=\bm{Q}\bm{W}^Q_h$, $\bm{K}_h=\bm{K}\bm{W}^K_h$, $\bm{V}_h=\bm{V}\bm{W}^V_h$, $H$ and $d_H$ denote the number and dimension of attention heads, respectively.  
where $\bm{W}^Q_h, \bm{W}^K_h, \bm{W}^V_h \in \mathbb{R}^{d_E \times d_H}$, and $d_H$ denotes the dimension of attention heads. For clarity, we write Eq.~\ref{eq-encoder-multi-head} in the form of vectors. The key point lies in that we incorporate the semantic relations between nodes into the \emph{query vector} ($\bm{q}_{n_j}$) and \emph{key vector} ($\bm{k}_{n_k}$):
\begin{eqnarray}\label{eq-encoder-relation}
\bm{q}_{n_j} &=& \bm{v}_{n_j} + \bm{v}_{n_j \to n_k}, \\
\bm{k}_{n_k} &=& \bm{v}_{n_k} + \bm{v}_{n_k \to n_j}, \nonumber
\end{eqnarray}
where $\bm{v}_{n_j \to n_k}$ and $\bm{v}_{n_k \to n_j}$ are encodings for the shortest relation path $n_j \to n_k$ and $n_k \to n_j$ between nodes $n_j$ and $n_k$, which are learned by summing the embeddings of the relations in the path. 

\ignore{Formally, we represent $j \to k = \langle z_0 \to z_1, z_1 \to z_2, \cdots ,z_n \to z_{n+1} \rangle$, where $z_0=j$, $z_{n+1}=k$ and $z_{1:n}$ are relay nodes. In this work, we compute the path encoding $\bm{v}_{j \to k}$ via:

\begin{equation}\label{eq-relation-sum}
\bm{v}_{j \to k} = \bm{v}_{z_0 \to z_1} + \cdots + \bm{v}_{z_n \to z_{n+1}}.
\end{equation}
where $\bm{v}_{z_p \to z_q}$ is the embedding of one-hop relation $z_p \to z_q$. 
}

Finally, we employ a residual connection and fully connected feed-forward network (FFN):
\begin{eqnarray}\label{eq-encoder-ffn}
\bm{\tilde{v}}_{n_j} &=& \bm{\hat{v}}_{n_j} + \text{FFN}(\bm{\hat{v}}_{n_j}), \\
\text{FFN}(\bm{x}) &=& \text{max}(0, \bm{x}^\top \bm{W}_1 + \bm{b}_1)\bm{W}_2 + \bm{b}_2, \nonumber
\end{eqnarray}
where $\bm{W}_1, \bm{W}_2, \bm{b}_1$ and $\bm{b}_2$ are trainable parameters, and FFN is a linear network with \texttt{gelu} activation.
%We can stack multiple blocks of the above architecture. 
The attention block, \ie MHA and FFN, can be stacked multiple times. 

\subsubsection{Subgraph Encoder}
\label{sec-two-level-attention}

Since a sentence plan corresponds to a connected subgraph from HKG, it naturally enforces the local coherence of its nodes. 
The major difficulty lies in how to enhance global coherence in text planning.  
Let $\bm{v}_{g_{j}} \in \mathbb{R}^{d_{E}}$ denote the embedding of the current subgraph $g_{j}$, initialized by an average pooling of the embeddings of all nodes in $g_{j}$ as $\bm{v}_{g_{j}} = \text{AvgPooling}(\bm{\tilde{v}}_{n_k})$, where $\bm{\tilde{v}}_{n_k}$ is the embedding of node $n_k$ in $g_{j}$ learned with Graph Transformer in Section~\ref{sec-graph-encoding}. Specifically, at the first step, the subgraph $g_0$ is initialized as a \textsc{Start} graph without any nodes.

As shown in Figure~\ref{fig-model}(d), the nodes in previous subgraphs have closed semantic correlations with the nodes in subsequent subgraphs. Therefore, we introduce two kinds of multi-head attention to enhance global coherence for subgraph representations. 

\paratitle{Subgraph-level Attention}. To make content globally coherent, the basic idea is to refer to previous subgraphs when learning the embedding of current subgraph. For this purpose, we propose a subgraph-level multi-head attention
 and obtain a subgraph-enhanced subgraph embedding $\bm{v}^G_{g_j}$ as:
\begin{equation}\label{eq-subgraph-level-attention}
\bm{v}^G_{g_{j}} = \underset{k < j}{\text{MHA}}(\bm{v}_{g_{j}}, \bm{v}_{g_k}, \bm{v}_{g_k}), 
\end{equation}
where $\bm{v}_{g_k}$ denotes the embeddings of previous subgraphs $g_1, \cdots, g_{j-1}$. In the subgraph-level multi-head attention mechanism, the embeddings for previous subgraphs are considered as key and value vectors, and the embedding of the current subgraph acts as the query vector. In this way, it incorporates the information of previous subgraphs for encoding the current subgraph. 

\paratitle{Node-level Attention}. Subgraph-level attention cannot directly reflect the fine-grained entity correlations between different subgraphs. Hence, we further propose to use node-level multi-head attention by considering the effect of nodes from \emph{previous subgraphs}.
The node-enhanced subgraph embedding $\bm{v}^N_{g_j}$ is given as:
\begin{equation}\label{eq-node-level-attention}
\bm{v}^N_{g_j} = \underset{n_z \in g_{k}, \forall k < j}{\text{MHA}}(\bm{v}^G_{g_j}, \bm{\tilde{v}}_{n_z}, \bm{\tilde{v}}_{n_z}),
\end{equation}
where $\bm{\tilde{v}}_{n_z}$ is the learned  embedding of node $n_z$ in previous subgraphs $g_1, \cdots, g_{j-1}$ computed as Eq.~\ref{eq-encoder-ffn}.
Similar to Eq.~\ref{eq-subgraph-level-attention}, the node embeddings in previous subgraphs are considered as key and value vectors, and the embedding of the current subgraph acts as the query vector. 
Hence, the node information of previous subgraphs has been incorporated for encoding the current subgraph.

With the two kinds of multi-head attention,  we have enhanced the global coherence by capturing inter-sentence correlations, since 
the information of previous subgraphs and their nodes can be injected into current subgraph. %In this way, we enhance global coherence by capturing inter-sentence correlations.
 Finally, we also apply a residual connection and fully connected feed-forward network (FFN) to the node-enhanced subgraph embedding $\bm{v}^N_{g_{j}}$ (similar to Eq.~\ref{eq-encoder-ffn}) and obtain the final subgraph embedding $\bm{\tilde{v}}_{g_{j}}$.

\subsubsection{Subgraph Decoder}

%Through the two-level multi-head attention, we can collect the history information from previous subgraphs and nodes in a hierarchical, top-down process, which is useful for predicting the next subgraph. 
After obtaining the final embedding of current subgraph $\bm{\tilde{v}}_{g_{j}}$, we further utilize it to generate the next subgraph, \ie $g_{j+1}$. 
General graph generation is a challenging task in deep learning~\cite{TrivediFBZ19,TaheriGB19}.
While, our task has several important unique characteristics, making the generation task simpler. 

\begin{figure}[t]
	\centering
	\includegraphics[width=0.45\textwidth]{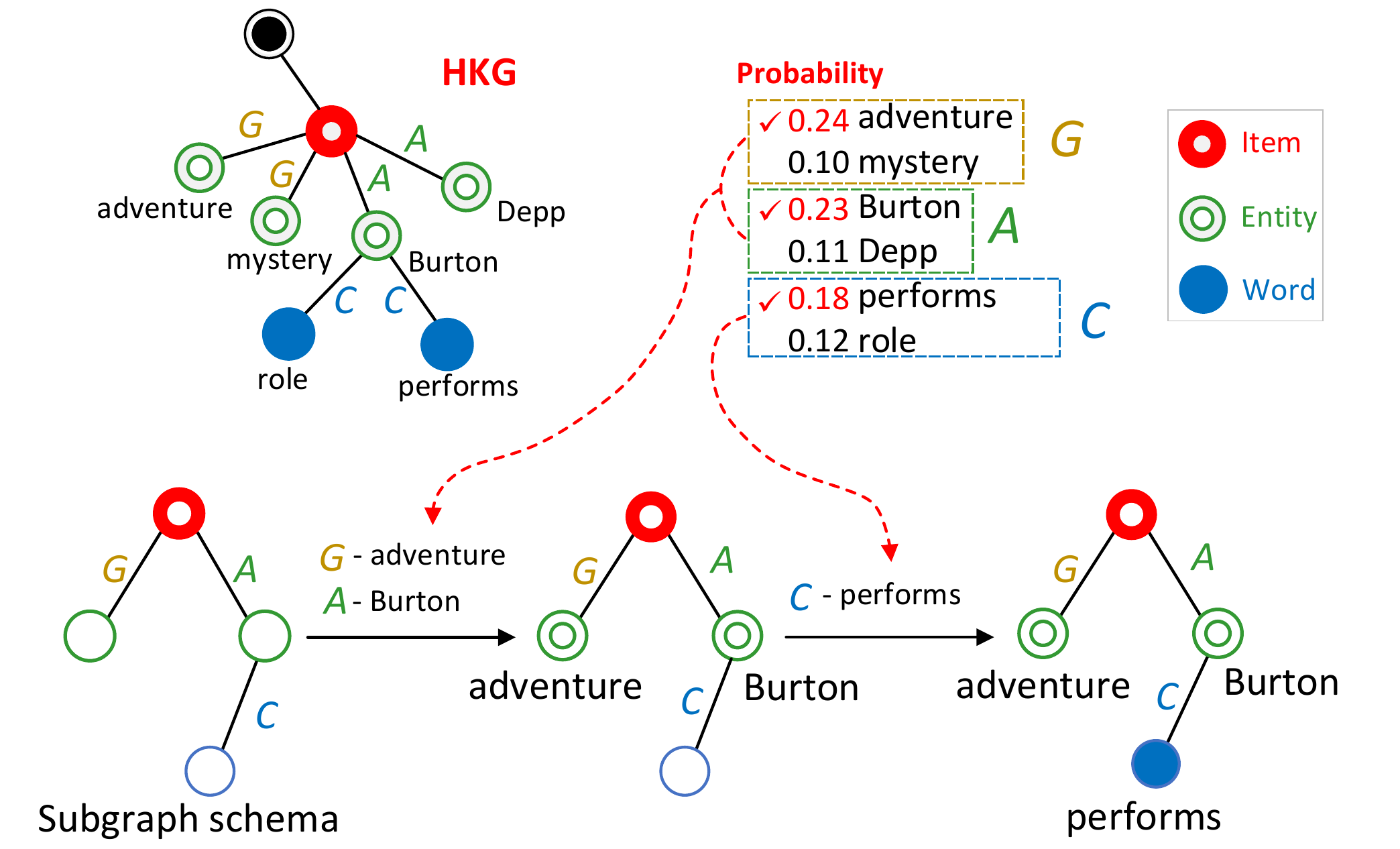} 
	\caption{An illustrative example for subgraph generation.}
	\label{fig-selection}
	%\vspace{-0.3cm}
\end{figure}

Recall that each subgraph $g_j$ is associated with a subgraph schema $s_j$ and a subgraph schema can be instantiated into different subgraphs (Section~\ref{sec-preliminary}). Thus, to generate a subgraph, we first generate the subgraph schema and then fill in the empty slots with entities or keywords. Usually, a review sentence usually contains only a few entities, and the number of frequent subgraph schemas in corpus is indeed small. 
We treat schema generation as a classification task over the frequent schema set, which is pre-extracted from training data. Once the schema has been determined, we utilize the relations in the schema as constraints and the entity (keyword) probability predicted by our model as selection criterion. Figure~\ref{fig-selection} presents an example for the process of subgraph generation. 

To enhance the personalized characteristics of subgraphs, following \emph{Attr2Seq}~\cite{ZhouLWDHX17}, we apply a standard attention mechanism~\cite{BahdanauCB14} on context information $\langle u,i,a \rangle$, and obtain a context vector $\bm{\tilde{c}}_j$ for encoding information of users, items and ratings.
\ignore{
\begin{eqnarray}\label{attn_score}
w_{j,k} &=& \frac{\exp(\tanh(\bm{W}_3 [\bm{\tilde{v}}_{g_j};\bm{v}_{c_k}]))}{\sum_{c_{k'} \in \{u,i,a\}} \exp (\tanh(\bm{W}_3 [\bm{\tilde{v}}_{g_j};\bm{v}_{c_{k'}}]))},  \\
\bm{\tilde{c}}_j &=& \sum_{c_{k} \in \{u,i,a\}} w_{j,k}\bm{v}_{c_k}, \nonumber 
\end{eqnarray}
where $\bm{W}_3$ is the trainable parameter, $\bm{v}_{c_k}$ is the embedding of context $c_k \in \{u,i,a\}$.} Finally, we stack the attention block, \ie subgraph- and node-level attention, by multiple times and compute the selection probabilities for a subgraph schema and an entity (or keyword) node as:
\begin{eqnarray}
\text{Pr}(s_{j+1} | g_1,...,g_j) &=& \text{softmax}(\bm{W}_4 [\bm{\tilde{v}}_{g_j}; \bm{\tilde{c}}_j] + \bm{b}_4),  \label{eq-schema-prob}\\
\text{Pr}(n_{j+1} | g_1,...,g_j) &=& \text{softmax}(\bm{W}_5 [\bm{\tilde{v}}_{g_j}; \bm{\tilde{c}}_j] + \bm{b}_5),  \label{eq-node-prob}
\end{eqnarray} 
where $\bm{W}_4, \bm{W}_5, \bm{b}_4$ and $\bm{b}_5$ are trainable parameters, and $s_{j+1}$ and $n_{j+1}$ denote a subgraph schema and an entity (or keyword), respectively. 
In practice, we select the most possible subgraph schema according to Eq.~\ref{eq-schema-prob}. Then, we collect all the entities that satisfy the requirement of the subgraph schema. Finally, each empty slot is filled in with the most probable node according to Eq.~\ref{eq-node-prob}. 
Although there might be other combinatorial optimization strategies, our method empirically works well and is more efficient. 
%More detailed description of subgraph generation can be referred to Appendix B.1. 

\subsection{Sentence Realization}
Given the inferred subgraph $g_{j}$, we next study how to generate the words of the $j$-th sentence, \ie $\langle w_{j,1},\cdots,w_{j,t},\cdots,w_{j,n_j} \rangle$. 
%We  start with a base sentence decoder by using GPT-2, and then extend it by introducing a supervised copy mechanism to incorporate entity information from the subgraph.
%The sentence generation module also uses the encoder-decoder architecture. The  graph encoders are similar to those in section~\ref{graph-based-aspect}. %Here we only focus on the decoding process of a single sentence. 

\subsubsection{Base Sentence Decoder}
\label{sec-base-sentence-decoder}

The base sentence generation module adopts Transformer decoder in GPT-2~\cite{radford2019language} by stacking  multiple self-attention blocks (similar to Eq.~\ref{eq-encoder-multi-head}\textasciitilde Eq.~\ref{eq-encoder-ffn}). Based on GPT-2, we can obtain the embedding $\bm{\tilde{v}}_{w_{j,t}} \in \mathbb{R}^{d_W}$ for the current word $w_{j,t}$ in the $j$-th sentence, where $d_W$ denotes the embedding size. Also, similar to Eq.~\ref{eq-schema-prob}-\ref{eq-node-prob}, we follow \emph{Attr2Seq}~\cite{ZhouLWDHX17} to encode 
context information $\langle u,i,a \rangle$ into a context vector $\bm{\tilde{c}}_{j,t}$ with attention mechanism.
%A slight difference is we also apply the standard attention mechanism on context information $\langle u,i,a \rangle$ and obtain context vector $\bm{\tilde{c}}_{j,t}$. 
%After stacking the self-attention block $L$ times, 
%
We generate the next word via a softmax probability function:
% with $\bm{\tilde{v}}_{w_{j,t-1}}$:
\begin{equation}\label{eq-base-prob}
\text{Pr}_{1}(w_{j,t+1}|w_{j,1},...,w_{j,t}) = \text{softmax}(\bm{W}_6 [\bm{\tilde{v}}_{w_{j,t}}; \bm{\tilde{c}}_{j,t}] + \bm{b}_6),
\end{equation}
where $\bm{W}_6$ and $\bm{b}_6$ are trainable parameters.
%We set the initial hidden vector for the $j$-th sentence as the last embedding of the previous sentence: $\bm{h}_{j,0}^S=\bm{h}_{j-1,n'_{j-1}}^S$. Specifically, we have $\bm{h}_{1,0}^S=\text{MLP}([\bm{v}_u ; \bm{v}_i ; \bm{v}_s])$ for initialization.

\subsubsection{Supervised Copy Mechanism} 
\label{sec-supervised-copy}

To verbalize KG subgraphs, we introduce a supervised copy mechanism that copies nodes from the subgraph.
The predictive probability  of a word $w$ can be decomposed into two parts, either generating a word from the vocabulary or copying a node from the subgraph:
\begin{eqnarray}\label{eq-sum-prob}
&&\text{Pr}(w_{j,t+1}=w|w_{j,1},...,w_{j,t}, g_j) \\
&=& \lambda_{j,t} \cdot \text{Pr}_{1}(w|w_{j,1},...,w_{j,t}) + (1-\lambda_{j,t}) \cdot \text{Pr}_{2}(w|g_j),\nonumber
\end{eqnarray} 
where $\text{Pr}_{1}(w|w_{j,1},...,w_{j,t})$ is the generative probability from the base sentence decoder  (Eq.~\ref{eq-base-prob}), and 
$\text{Pr}_{2}(w|g_j)$ is the copy probability  defined as:
\begin{equation}\label{eq-copy-prob}
\text{Pr}_{2}(w|g_j) = \frac{\exp(\tanh(\bm{W}_7 [\bm{\tilde{v}}_{w_{j,t}}; \bm{\tilde{c}}_{j,t}; \bm{\tilde{v}}_w]))}{\sum_{w' \in g_j}\exp(\tanh(\bm{W}_7 [\bm{\tilde{v}}_{w_{j,t}}; \bm{\tilde{c}}_{j,t}; \bm{\tilde{v}}_{w'}]))},
\end{equation}
where $\bm{W}_7$ is the trainable parameter and $\bm{\tilde{v}}_w$ is the embedding of an entity or a word node $w$ in the current subgraph $g_j$.
Note that we only copy entities or keywords from the predicted subgraph, which dramatically reduces the candidate set. Since subgraph generation has already considered local and global coherence, our candidate set is more meaningful and coherent. In Eq.~\ref{eq-sum-prob}, we use a dynamically learned coefficient $\lambda_{j,t}$ to control the combination between the two parts as:
\begin{equation}\label{eq-copy-alpha}
\lambda_{j,t} = \sigma (\bm{w}^\top_{gen}[\bm{\tilde{v}}_{w_{j,t}};\bm{\tilde{c}}_{j,t}]+ b_{gen}),
\end{equation} 
where $\bm{w}_{gen}$ and $b_{gen}$ are trainable parameters. For each word, we add a binary indicator $d_{j,t}$ (0 for \emph{copy} and 1 for \emph{generate}) to provide a supervised signal for the generation and copy. In addition to the word prediction loss, we incorporate a supervised indicator loss with the 
binary cross entropy:
\begin{equation}\label{eq-super-loss}
\mathcal{L}_{si} = -\sum\limits_{j,t} d_{j,t} \log (\lambda_{j,t}) - (1-d_{j,t}) \log (1-\lambda_{j,t}).
\end{equation} 
Different from traditional copy mechanism, we utilize the loss in Eq.~\ref{eq-super-loss} to explicitly guide the switch between copy or generation during decoding, which can further enhance the local coherence via copying threading keywords from subgraphs. 

\subsection{Discussion and Learning}
In this part, we present the model discussion and optimization. 

\paratitle{Coherence}.  For local coherence, we utilize KG subgraphs as sentence plans, since KG subgraphs are tightly associated semantic structures, which naturally enforce the intra-sentence correlations of  entities. Supervised copy mechanism (Section~\ref{sec-supervised-copy}) further connects entities in sentences with the copied threading words from subgraphs. For global coherence, we utilize both subgraph- and node-level attention (Section~\ref{sec-two-level-attention}) to enhance inter-sentence correlations of entities.
Note that not all sentences include entity mentions, we set up a special sentence plan that directly calls the base decoder in Section~\ref{sec-base-sentence-decoder} without copy mechanism. To our knowledge, there are  seldom studies that  consider both local and global coherence in text generation models. By incorporating KG data, our model provides a principled text planning approach for enhancing the two kinds of coherence of the generated text. 

\paratitle{Personalization}.  Review generation requires to capture personalized user preference and  writing styles. We explicitly model personalization through the contextual embeddings of users, items, and ratings during the decoding of subgraphs and words in Eq.~\ref{eq-schema-prob}\textasciitilde \ref{eq-node-prob} and Eq.~\ref{eq-base-prob}\textasciitilde \ref{eq-copy-prob}, respectively. Another point is that HKG embedding (Section~\ref{sec-graph-encoding}) has involved user-item interactions, which can capture user preference over items and associated attributes. In particular, given a $\langle user, item \rangle$ pair, we construct the HKG by involving one-hop entities linked with the item from KG and the associated keywords for entities. Such a method naturally enforces the personalized preference over item attributes for a given user. 

\paratitle{Optimization}. In our model, there are two sets of trainable parameters in subgraph generation and sentence generation, denoted by $\Theta^{(s)}$ and $\Theta^{(w)}$, respectively. 
First, we optimize $\Theta^{(s)}$ according to the predictive loss for subgraph schemas and nodes based on  cross entropy loss using Eq.~\ref{eq-schema-prob} and Eq.~\ref{eq-node-prob}.
And then, we optimize $\Theta^{(w)}$ according to the indicator loss in Eq.~\ref{eq-super-loss} and word prediction loss that sums negative likelihood of individual words using Eq.~\ref{eq-sum-prob}.
We incrementally train the two parts, and fine-tune the shared or dependent parameters in different modules.
For training, we directly use the real subgraphs and sentences to optimize the model parameters with Adam optimizer~\cite{KingmaB14}. The same learning rate schedule in \cite{VaswaniSPUJGKP17} is adopted in our training. In order to avoid overfitting, we adopt the dropout strategy with a ratio of 0.2. 
During inference, we apply our model in a pipeline way: we first infer the subgraph sequence, then predict the sentences using inferred subgraphs. For sentence generation, we apply the beam search method with a beam size of 8.
We set the maximum generation lengths for subgraph and sentence sequence to be 5 and 50, respectively. %We present the detailed training algorithms in Appendix B.3.

\section{Experiments}

In this section, we conduct the evaluation experiments for our approach on the review generation task. We first set up the experiments, and then report the results and detailed analysis.

\subsection{Experimental Setup}
\subsubsection{Construction of the Datasets} We use three datasets from different domains for evaluation,  including \textsc{Amazon} Electronic, Book datasets~\cite{HeM16}, and \textsc{IMDb} Movie dataset~\cite{imdb}.  
We remove users and items occurring fewer than five times, discard reviews containing more than 100 tokens
and only keep the top frequent 30,000 words in vocabulary for the three datasets.
All the text is processed with the procedures of lowercase and tokenization using NLTK.
In order to obtain KG information for these items, we adopt the public KB4Rec~\cite{zhao2019kb4rec} dataset to construct the aligned linkage between Freebase~\cite{freebase} (March 2015 version) entities and online items from the three domains. %Note that not all the items can be aligned to Freebase entities, and we only keep the data of the aligned items.
Starting with the aligned items as seeds, we include their one-hop neighbors from Freebase as our KG data.
We keep the reverse relations and remove the triples with non-Freebase strings. 
Note that we only retain the items linked with Freebase entities in our datasets.
The statistics of three datasets after preprocessing are summarized in Table~\ref{tab-data}.
Furthermore, for each domain, we randomly split it into training, validation  and test sets with a ratio of 8:1:1.
To construct the entity-word links in HKG, we employ the Stanford NER package to identify entity mentions in review text, and extract aspect words by following~\cite{NiM18}. 
We select 489, 442 and 440 aspect words, frequently co-occurring with entity mentions in review sentences, for the three domains, respectively. 
The user-item links in HKG can be constructed according to user-item interactions in review datasets. 
Finally, we extract the top 30, 30, and 35 frequent subgraph schemas using gSpan algorithm~\cite{YanH02} for the three domains, respectively.

\begin{table}[tbp]
	\centering\small
	\caption{Statistics of our datasets after preprocessing.}
	\begin{tabular}{c|l|r r r}
		\toprule[1pt]
		\multicolumn{2}{c|}{\textbf{Dataset}}& \textbf{Electronic} & \textbf{Book} & \textbf{Movie}\\
		\midrule[0.7pt]
		\multirow{3}{*}{\tabincell{c}{Review}}&\#Users 	&	50,473&	71,156&	47,096\\
		&\#Items 	&12,352&	25,045&	21,125\\
		&\#Reviews 	&	221,722&	853,427&	1,152,925\\
		\midrule[0.7pt]
		\multirow{3}{*}{\tabincell{c}{Knowledge\\Graph}}&\#Entities	&30,310	&105,834	&247,126	\\
		&\#Relations	&30	&12	&16	\\
		&\#Triplets 	&129,254	&300,416	&1,405,348	\\
		\bottomrule[1pt]
	\end{tabular}%
	\label{tab-data}%
\end{table}

\subsubsection{Baseline Methods} We consider the following baselines as comparison: 

\textbullet~\emph{Attr2Seq}~\citep{ZhouLWDHX17}: It adopts an attention-enhanced attribute-to-sequence framework to generate reviews with input attributes.
% In order to address the problem of long-range dependency, the context information are attended through a gating mechanism.

\textbullet~\emph{A2S+KG}: We incorporate the KG embeddings of items as additional inputs into \emph{Attr2Seq}.

\textbullet~\emph{ExpanNet}~\citep{NiM18}: It adopts an encoder-decoder architecture to generate personalized reviews by introducing aspect words.
% information (\eg aspect words). 

\textbullet~\emph{A-R2S}~\citep{NiLM19}: It employs a reference-based Seq2Seq model with aspect-planning in order to cover different aspects.

\textbullet~\emph{A-R2S+KG}: We incorporate the KG entities of items as external inputs into \emph{A-R2S}.

\textbullet~\emph{SeqGAN}~\citep{YuZWY17}: : It regards the generative model as a stochastic parameterized policy and uses Monte Carlo search to approximate the state-action value. The discriminator is a binary classifier to evaluate the sequence and guide learning process of the generator.

\textbullet~\emph{LeakGAN}~\citep{GuoLCZYW18}: It is designed for long text generation through the leaked mechanism. The generator is built upon a hierarchical reinforcement learning architecture and the discriminator is a CNN-based feature extractor. 

\textbullet~\emph{ACF}~\citep{LiZWS19}: It decomposes the review generation process into three different stages by designing an aspect-aware coarse-to-fine generation model. The aspect semantics and syntactic characteristics are considered in the process.

\textbullet~\emph{KCGNN}~\citep{li2020knowledge}: It proposes a KG-enhanced review generation model based on capsule graph neural network for capturing user preference at both aspect and word levels.

\textbullet~\emph{PHVM}~\citep{ShaoHWXZ19}: It adopts a planning-based hierarchical variational model to capture the inter-sentence coherence of texts.

Among these baselines, \emph{Attr2Seq}, \emph{ExpanNet}, \emph{A-R2S}, \emph{ACF} and \emph{KCGNN} are five recently proposed review generation models; \emph{SeqGAN} and \emph{LeakGAN} are GAN-based text generation models; \emph{A2S+KG} and \emph{A-R2S+KG} incorporate the pre-trained KG item embeddings from DistMult~\cite{YangYHGD14a} and KG entities of items, respectively; \emph{PHVM} is the state-of-the-art text planning model. We implement it by transferring KG into a list of $\langle relation, entity \rangle$ pairs (\eg $\langle actor, Burton \rangle$) about items (\eg movie $Sleepy$) as inputs. We employ validation set to optimize the parameters in each method. %Besides, due to similar performance, we omit results of several other studies~\cite{TangYCZM16,CatherineC18,LiT19}. 
%Besides, several other studies, including gC2S~\cite{TangYCZM16}, TransNets~\cite{CatherineC18} and RevGAN~\cite{LiT19}, can be considered as baselines. However, these candidates at least resemble to one of the compared baselines, having similar performance. Furthermore, we also incorporate KG data into \emph{ExpanNet} and \emph{ACF}, but achieve lower results than \emph{KCGNN} and \emph{PHVM}. We omit their results in comparison for simplicity. All the parameters are randomly initialized by sampling from a normal distribution with mean $0.0$ and standard deviation $0.02$. We employ validation set to optimize the parameters in each method. %To reproduce the results of our model, we report the parameter setting used throughout the experiments in Table~\ref{tab-parameters}.

\ignore{
\begin{table}
	\centering\small
	\caption{Parameter settings of  the two modules in our model.} 
	\begin{tabular}{c | l}
		\toprule[1pt]
		\textbf{Modules} & \textbf{Settings} \\
		\midrule[0.7pt]
		Subgraph & \tabincell{l}{$d_E=512$, FFN-embed.-size=$1024$, batch-size=$128$,\\
			Attention-heads=$8$, Attention-blocks=$6$,\\
			init.-learning-rate=$0.002$, Adam optimizer}\\
		\midrule[0.7pt]
		Sentence & \tabincell{l}{$d_W=512$, FFN-embed.-size=$1024$, batch-size=$64$,\\
			Attention-heads=$8$, Attention-blocks=$6$,\\
			init.-learning-rate=$0.002$, Adam optimizer}\\
		\bottomrule[1pt]
	\end{tabular}
	\label{tab-parameters}
\end{table}}

\begin{table*}[tb]
	\renewcommand\arraystretch{1.1}
	\small
	\begin{center}
		\caption{Performance comparisons of different methods for automatic review generation under three domains. ``*'' denotes the improvement is statistically significant compared with the best baseline (t-test with p-value $< 0.05$). ``-'' denotes this metric is not applicable to this method, since the generated text contains very few entity mentions.}
		\begin{tabular}{c|l | c c | c c c c c}
			\toprule[1pt]
			\multirow{2}[2]*{\textbf{Datasets}} & \multirow{2}[2]*{\textbf{Models}} & \multicolumn{2}{c|}{\textbf{Coherence}} & \multicolumn{5}{c}{\textbf{Generation}} \\
			\cmidrule[0.7pt]{3-9}
			& & \textbf{Sen-Sim} & \textbf{ECR} & \textbf{BLEU-1} & \textbf{BLEU-4} & \textbf{ROUGE-1} & \textbf{ROUGE-2} & \textbf{ROUGE-L} \\
			\midrule[0.7pt]
			\multirow{4}[6]{*}{\textsc{Electronic}}
			%& gC2S & 38.67 & 24.14 & 0.85 & 0.262 & 0.046 & 0.212 \\
			& Attr2Seq & 0.671 & - & 24.28 & 0.88 & 0.263 & 0.043 & 0.214 \\
			& A2S+KG & 0.652 & 1.63 & 25.62 & 0.93 & 0.271 & 0.049 & 0.223 \\
			& ExpanNet & 0.653 & - & 26.56 & 0.95 & 0.290 & 0.052 & 0.262 \\
			& A-R2S & 0.667 & - & 27.04 & 1.15 & 0.309 & 0.065 & 0.279 \\
			& A-R2S+KG & 0.650 & 7.01 & 29.28 & 1.69 & 0.322 & 0.067 & 0.288 \\
			%\cline{2-8}
			& SeqGAN & 0.625 & - & 25.18 & 0.84 & 0.265 & 0.043 & 0.220 \\
			& LeakGAN & 0.666 & - & 25.66 & 0.92 & 0.267 & 0.050 & 0.236 \\
			%
			%& TransNets & 34.21& 21.61 & 0.60 & 0.227 & 0.026 & 0.199 \\
			%
			%& RevGAN & 32.79 & 25.50 & 0.75 & 0.240 & 0.030 & 0.222 \\
			& ACF & 0.686 & - & 28.22 & 1.04 & 0.315 & 0.066 & 0.280 \\
			& KCGNN & 0.688 & 16.50 & \underline{29.88} & 1.83 & 0.323 & \underline{0.078} & 0.295 \\
			& PHVM & \underline{0.690} & \underline{17.56} & 29.40 & \underline{1.93} & \underline{0.325} & 0.072 & \underline{0.301} \\
			%\cline{2-8}
			%\cline{2-8}
			& CETP & \textbf{0.707*} & \textbf{20.85*} & \textbf{31.51*} & \textbf{3.12*} & \textbf{0.338*} & \textbf{0.084*} & \textbf{0.312*} \\
			\midrule[0.7pt]
			\multirow{4}[6]{*}{\textsc{Book}}
			%& gC2S & 30.58 & 25.87 & 1.03 & 0.265 & 0.044 & 0.217 \\
			& Attr2Seq & 0.677 & - & 26.93 & 1.14 & 0.259 & 0.047 & 0.223 \\
			& A2S+KG & 0.672 & 2.25 & 27.69 & 1.42 & 0.268 & 0.053 & 0.236 \\
			%& TransNets & 32.45 & 23.25 & 0.74 & 0.224 & 0.030 & 0.152 \\
			%\cline{2-8}
			%& RevGAN & 32.13 & 25.25 & 0.86 & 0.267 & 0.052 & 0.217 \\
			%& SeqGAN & 27.11 & 26.89 & 1.24 & 0.255 & 0.053 & 0.246 \\
			& ExpanNet & 0.708 & - & 26.52 & 1.49 & 0.301 & 0.054 & 0.271 \\
			& A-R2S & 0.695 & - & 28.34 & 1.82 & 0.318 & 0.075 & 0.283 \\
			& A-R2S+KG & 0.671 & 9.07 & 29.00 & 2.06 & 0.321 & 0.077 & 0.295 \\
			%\cline{2-8}
			& SeqGAN & 0.633 & - & 26.89 & 1.24 & 0.255 & 0.053 & 0.246 \\
			& LeakGAN & 0.663 & - & 28.79 & 1.94 & 0.274 & 0.060 & 0.285 \\
			& ACF & 0.715 & - & 28.96 & 2.11 & 0.317 & 0.068 & 0.291  \\
			& KCGNN & 0.733 & 16.66 & \underline{30.66} & \underline{3.08} & \underline{0.332} & 0.080 & 0.306 \\
			& PHVM & \underline{0.740} & \underline{18.79} & 29.33 & 2.46 & 0.319 & \underline{0.085} & \underline{0.307} \\
			%\cline{2-8}
			& CETP & \textbf{0.761*} & \textbf{23.89*} & \textbf{31.93*} & \textbf{3.89} & \textbf{0.341*} & \textbf{0.095*} & \textbf{0.317*} \\
			\midrule[0.7pt]
			\multirow{4}[6]{*}{\textsc{Movie}}
			%& gC2S & 34.12 & 26.17 & 1.09 & 0.272 & 0.047 & 0.215 \\
			& Attr2Seq & 0.629 & - & 26.57 & 1.55 & 0.271 & 0.050 & 0.222 \\
			& A2S+KG & 0.619 & 14.96 & 27.02 & 1.67 & 0.278 & 0.053 & 0.235 \\
			& ExpanNet & 0.651 & - & 27.93 & 2.00 & 0.310 & 0.063 & 0.266 \\
			& A-R2S & 0.649 & - & 29.01 & 2.12 & 0.314 & 0.074 & 0.306 \\
			& A-R2S+KG & 0.617 & 20.06 & 30.05 & 2.95 & 0.325 & 0.077 & 0.313 \\
			%& TransNets & 32.46 & 23.87 & 0.76 & 0.221 & 0.031 & 0.155 \\
			%\cline{2-8}
			%%& RevGAN & 28.22 & 24.68 & 0.91 & 0.260 & 0.054 & 0.198 \\
			%& SeqGAN & 24.53 & 27.07 & 1.63 & 0.274 & 0.052 & 0.221 \\
			%\cline{2-8}
			& SeqGAN & 0.641 & - & 27.07 & 1.63 & 0.274 & 0.052 & 0.221 \\
			& LeakGAN & 0.672 & - & 28.10 & 2.29 & 0.302 & 0.064 & 0.271 \\
			& ACF & 0.709 & - & 29.46 & 2.40 & 0.322 & 0.076 & 0.303  \\
			& KCGNN & 0.766 & 23.34 & \underline{31.39} & \underline{3.55} & \underline{0.341} & 0.096 & 0.327 \\
			& PHVM & \underline{0.770} & \underline{24.70} & 30.29 & 3.02 & 0.331 & \underline{0.098} & \underline{0.328} \\
			%\cline{2-8}
			& CETP & \textbf{0.794*} & \textbf{28.96*} & \textbf{32.37*} & \textbf{4.39*} & \textbf{0.353*} & \textbf{0.114*} & \textbf{0.345*} \\
			\bottomrule[1pt]
		\end{tabular}
		\label{tab:main-results}
	\end{center}
	\vspace{-0.04cm}
\end{table*}

\paratitle{Evaluation Metrics}. To evaluate the performance of review generation, 
we adopt two automatic \emph{generation metrics}, including BLEU-1/4 and ROUGE-1/2/L. 
BLEU~\cite{PapineniRWZ02} measures the ratios of the co-occurrences of $n$-grams between the generated and real reviews;
ROUGE~\cite{Lin04} counts the overlapping $n$-grams between generated and real reviews.
Furthermore, to evaluate the coherence of generated reviews, we adopt two automatic \emph{coherence metrics}, including Sen-Sim proposed in~\cite{LapataB05} (measuring discourse coherence as an average cosine similarity between any two sentences from the discourse based on sentence embeddings from BERT~\cite{DevlinCLT19}) and Entity Co-occurrence Ratio~(abbreviated as \emph{ECR}, modified based on BLEU-2 and computing the ratio of co-occurrences of entity pairs between generated and real reviews). 
Compared with~\cite{LapataB05} which represents the sentence embedding as the mean of embeddings of words in the sentence, BERT adds a special token ``\texttt{[CLS]}'' as the first token of every sentence and the final representation of this token is used as the sentence embedding. We also try other models (\eg Word2Vec and ELMo) to acquire sentence embeddings. These models achieve similar results as BERT. %Detailed formulation and description about Sen-Sim and ECR are reported in the Appendix C.2.  

\subsection{Main Results}

%\subsubsection{Performance Comparison}

Table~\ref{tab:main-results} presents the results of different methods on the review generation task. 

First, by incorporating the KG embeddings or KG entities, A2S+KG and A-R2S+KG achieve better results for generation metrics while worse results for coherence metrics than original Attr2Seq and A-R2S. This observation implies that, although KG data is useful for text generation, we should more carefully incorporate it considering the text coherence. Besides, among the GAN-based methods, LeakGAN gives better results for both kinds of metrics than SeqGAN. A major reason is that LeakGAN is specially designed for generating long text, while SeqGAN may not be effective in capturing long-range semantic dependency in text generation.

Second, ACF, KCGNN and PHVM overall outperform most of baselines for both kinds of metrics. 
%Specifically, 
ACF adopts a coarse-to-fine  three-stage generation process by considering aspect semantics and syntactic patterns, and KCGNN is a KG-enhanced generation model for capturing user preference on KG attributes. It shows that both aspect semantics and KG information are helpful for review generation, especially KG information. As the most relevant comparison with our model,  the recent proposed PHVM yields the best performance among all baselines. It introduces KG entities and verbalizes coherent sentences conditioned on attribute-level planning (a sequence of entity groups). 

\ignore{Second, LeakGAN and ACF outperform most of baselines for generation and coherence metrics due to the special long text modeling. Specifically, ACF adopts a more powerful three-stage generation process by considering aspect semantics and syntactic patterns.
Among all baselines, the most competitive and direct comparison with our model is PHVM, which realizes coherent sentences conditioned on attribute-level planning (a sequence of item groups).
}
%GAN-based methods work better than the above baselines, especially LeakGAN. LeakGAN is specially designed for generating long text, and we adapt it to our task by incorporating context information. 

%Third, ExpansionNet and Coarse-to-Fine models perform better than the above all baselines and Coarse-to-Fine improves even better. The major reason is that ExpansionNet incorporates external knowledge such as review summaries, product titles and aspect keywords, and Coarse-to-Fine decompose the difficult generation process into three simple and controllable modules, considering the aspect semantics and syntactics. 

Finally, we compare the proposed CETP with the baseline methods. It is clear to see that CETP performs better than all the baselines by a large margin. %Our model effectively utilizes KG information for developing a KG-enhanced two-stage generation process. 
The major difference between our model and baselines lies in that we design a text planning mechanism based on KG subgraphs in the generation process, thus simultaneously improving the global and local coherence of texts.
PHVM lacks the modeling of multi-grained correlations between entity groups, and also neglects the  intrinsic  structure of an entity group.
%PHVM does not consider the  associations within each attribute group, and fails in modeling intrinsic coherence for a group.  
While, other baselines do not explicitly model the coherence of text or incorporate external KG data.
% Overall, the three datasets show the similar findings, especially in movie dataset. A possible reason is that the movie dataset has more dense KG data. 

\begin{table}[t]
	\centering
	\caption{Ablation analysis on \textsc{Movie} dataset.} 
	\begin{tabular}{ l c c c  }
		\toprule[1pt]
		\textbf{Models} & \textbf{ECR} & \textbf{BLEU-1} & \textbf{ROUGE-1} \\
		\midrule[0.7pt]
		CETP & 28.96 & 32.37 & 0.342  \\
		\midrule[0.7pt]
		%w/o KG & 18.78 & 29.29 & 0.315 \\
		w/o HKG, w KG & 28.91 & 31.93 & 0.335  \\
		%w/o Subgraph & 22.34 & 30.29 & 0.320  \\
		w/o SA & 25.60 & 30.96 & 0.317 \\
		w/o NA & 26.00 & 31.34 & 0.330\\
		w/o Copy & 25.20 & 31.03 & 0.326  \\
		\bottomrule[1pt]
	\end{tabular}
	\label{tab:ablation-results}
\end{table}

\subsection{Detailed Analysis}
Next, we construct detailed analysis experiments on our model.
We only report the results on \textsc{Movie} dataset due to similar findings in three datasets. We select the two best baselines \emph{KCGNN} and \emph{PHVM} as comparisons.

\subsubsection{Ablation Analysis} Our model has three novel designs: HKG incorporation, subgraph-based text planning and supervised copy mechanism. 
Table~\ref{tab:ablation-results} shows the results if we ablate these designs. Here, we consider four variants for comparison: (1) \emph{w/o HKG, w KG} keeping original KG links while removes the interaction and co-occurrence links in HKG; (2) \emph{w/o SA} removing subgraph-level attention (Eq.~\ref{eq-subgraph-level-attention}) in subgraph-based text planning; (3) \emph{w/o NA} removing node-level attention (Eq.~\ref{eq-node-level-attention}) in subgraph-based text planning; (4) \emph{w/o Copy} removing the supervised copy mechanism during generating words (Section~\ref{sec-supervised-copy}).

In Table~\ref{tab:ablation-results}, we can see that removing user and word nodes (including the associated links) gives a worse result than CETP, which shows that user-item interaction and entity-word co-occurrence are useful to review generation in terms of capturing user preference and entity-keyword association.
Second, variants dropping the subgraph- and node-level attention are worse than CETP, especially dropping the subgraph-level attention.
This shows that our model benefits from the subgraph-based text planning, which improves the process of content selection, arrangement, and order.
Finally, removing the supervised copy mechanism also greatly declines the final performance of our model.
In our model, the supervised copy mechanism explicitly guides the switch between generation and copy by selecting highly related entities or words from the planned subgraph, which has a significant effect on the final coherence performance. 
   
%$\bullet$ \emph{w/o Graph}: the variant removes the graph without any knowledge and keywords information.

%$\bullet$ \emph{w/o KG}: the variant removes the KG data from the our heterogeneous graph, but keep the other nodes and links. %We attach word nodes to the item nodes. 

%$\bullet$ \emph{w/o Capsule}: the variant replaces the Capsule GNN with a conventional R-GCN method. 

%$\bullet$ \emph{w/o Copy}: the variant removes the copy mechanism when generating sentences. 

% that it is important to adopt suitable ways to utilize the knowledge graph when predicting the aspect sequence and review sentences. 

\subsubsection{Human Evaluation} Above, we have performed automatic evaluation experiments for our model and baselines. For text generation models, it is important
to construct human evaluation for further effectiveness verification. 

Following previous work~\cite{LiZWS19,NiLM19}, we also conduct human evaluation on the generated reviews. 
We randomly choose 200 samples from test set. 
A sample contains the input contexts (\ie user, item and rating), and the texts generated by different models.
Three experienced e-commerce users were asked to score the texts with respect to four dimensions of coherence, relevance, fluency and informativeness.
\emph{Coherence} evaluates how content is coherent considering both intra- and inter-sentence correlation~\cite{ShaoHWXZ19}.
\emph{Relevance} measures how relevant the generated review is according to the input contexts.
\emph{Fluency} measures how likely the generated review is produced by human.
\emph{Informativeness} means that how much the generated text provides specific or different information.

The scoring mechanism adopts a 5-point Likert scale~\cite{likert1932technique}, ranging from 1-point~(``very terrible'') to 5-point~(``very satisfying'').
% which 5-point means ``very satisfying'', and 1-point means ``very terrible''. 
We further average the three scores from the three human judges over the 200 inputs for each method. 
The results in Table \ref{tab:human-results} show that CETP produces more coherent texts, which further verifies the effectiveness of the subgraph-based text planning. It is also worth noting that CETP performs better in terms of fluency, since KG subgraphs enforce more fluent and logical expressions.
The informativeness of CETP is slightly worse than PHVM. It is possibly because PHVM applies a more greedy strategy to copy entities from KG while our model adopts a more conservative strategy to incorporate highly relevant KG entities.
The Cohen's kappa coefficients for the four factors are 0.78, 0.71, 0.75 and 0.69, respectively, indicating a high agreement between the three human judges.

\begin{table}[t]
	\centering
	\caption{Human evaluation on  \textsc{Movie} dataset.} 
	\begin{tabular}{l c c c c }
		\toprule[1pt]
		\textbf{Metrics} & \textbf{Gold} & \textbf{CETP} &  \textbf{KCGNN} & \textbf{PHVM} \\
		\midrule[0.7pt]
		Coherence & 4.22 & \underline{3.51} & 3.10 & 3.18 \\
		Relevance & 4.22 & \underline{3.42} & 3.34 & 3.33 \\
		Fluency & 4.54 & \underline{3.49} & 3.15 & 3.08  \\
		Informativeness & 4.33 & 2.97 & 2.95 & \underline{3.03} \\
		\bottomrule[1pt]
	\end{tabular}
	\label{tab:human-results}
\end{table}

\begin{figure}[h]
	\centering
	\subfigure[Tuning the amount of KG data.]{\label{fig-movie-varing-kg}
		\centering
		\includegraphics[width=0.22\textwidth]{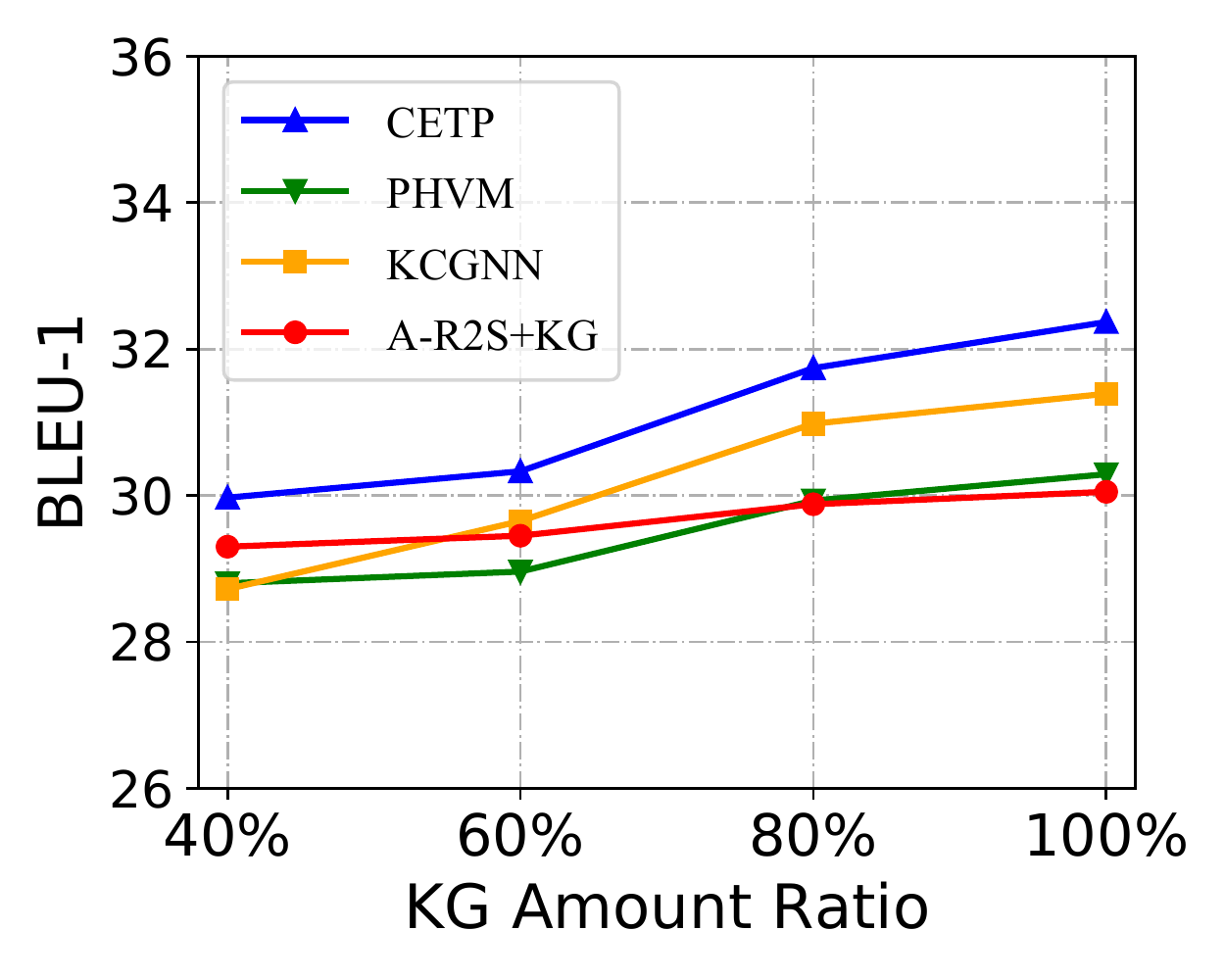}
	}
	\subfigure[Tuning the KG embedding size.]{\label{fig-movie-varing-es}
		\centering
		\includegraphics[width=0.22\textwidth]{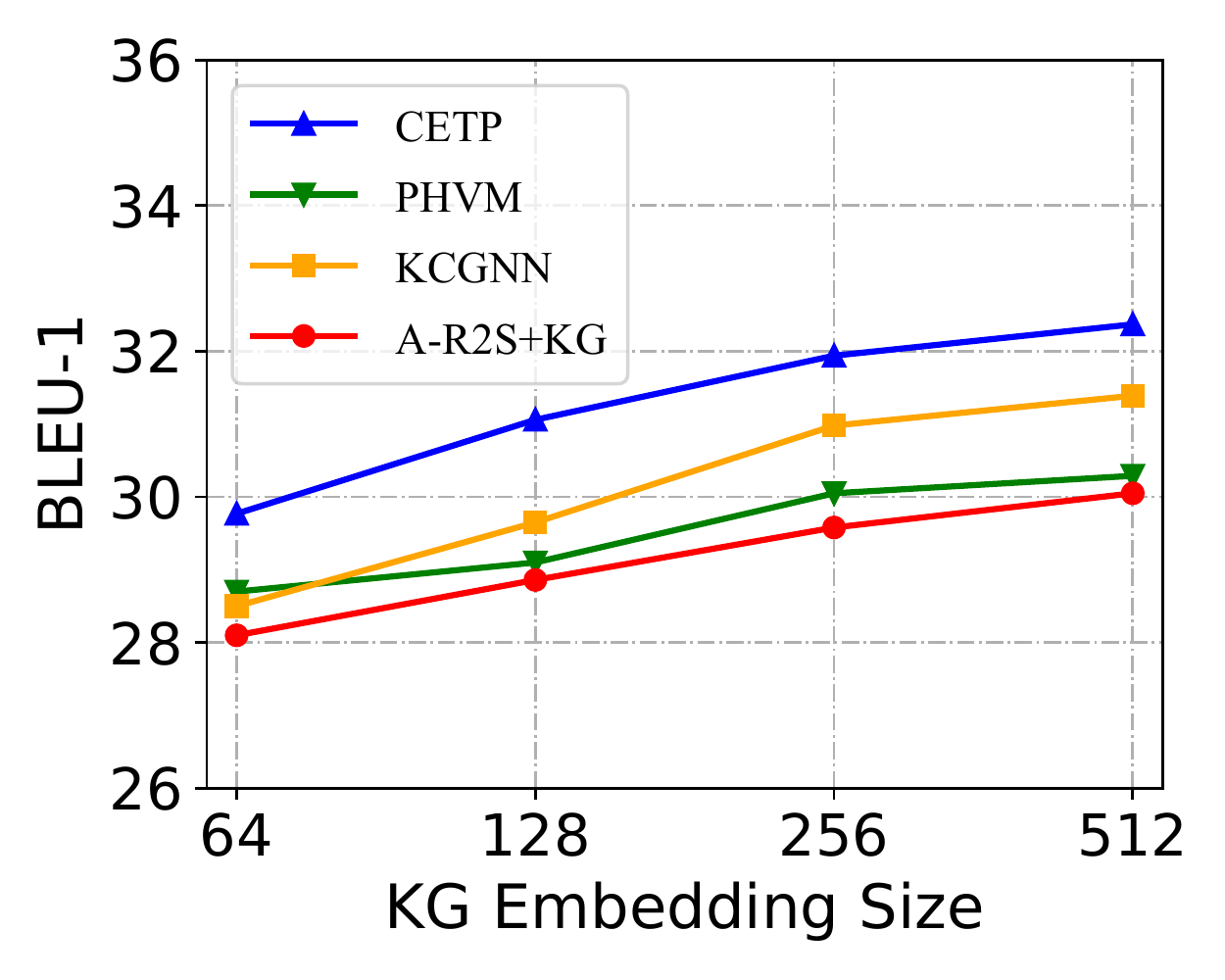}
	}
	\centering
	\caption{Performance tuning  on \textsc{Movie} dataset.}
	\label{fig-parameter-tuning-movie}
\end{figure}

\begin{figure*}[t]
	\centering
	\includegraphics[width=0.98\textwidth]{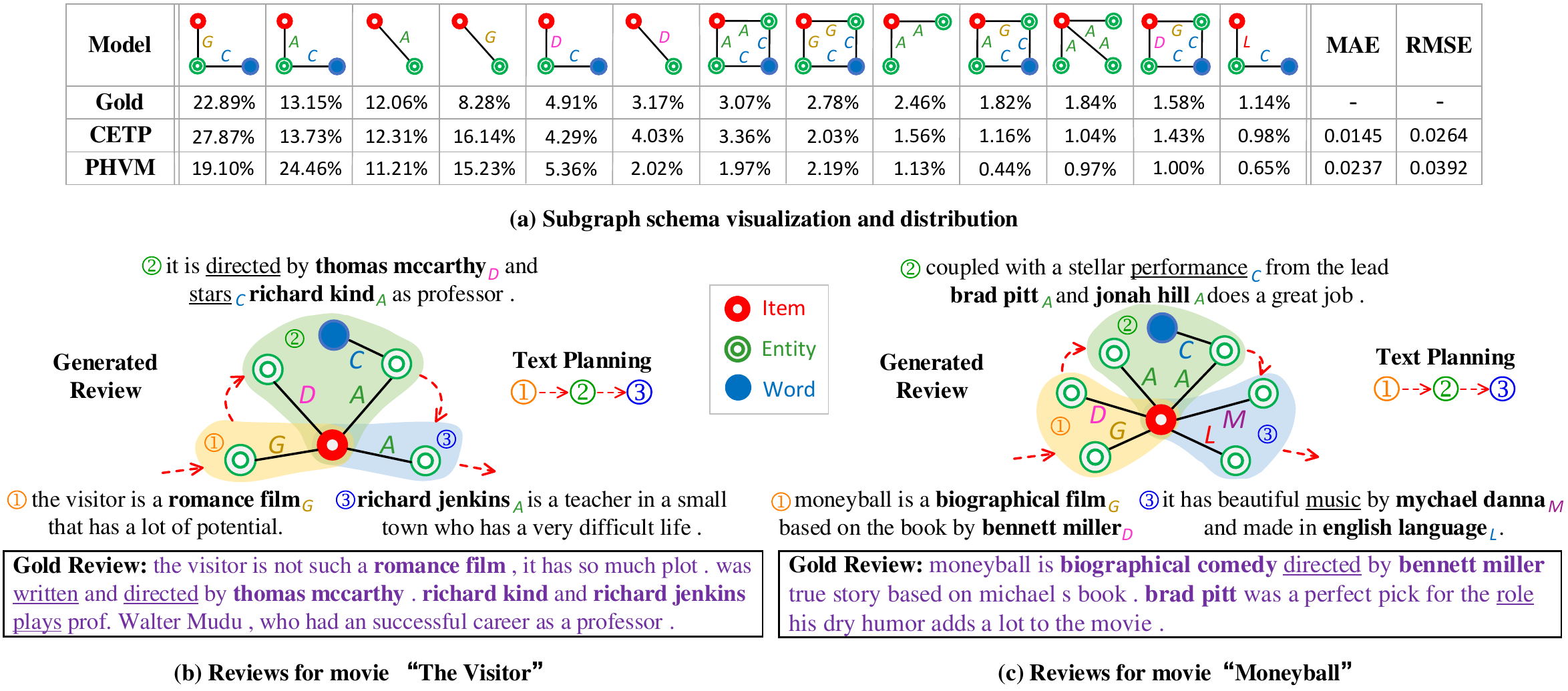} 
	\caption{Subgraph schema visualization and sample reviews generated by CETP on \textsc{Movie} dataset. The two reviews are about the movies \emph{``The Visitor''} and \emph{``Moneyball''} from the same user. The capital letters $A$, $G$, $D$, $L$, $M$ and $C$ denote the relations of \emph{actor}, \emph{genre}, \emph{director}, \emph{language}, \emph{music} and \emph{co-occurrence}, respectively.}
	\label{fig-case}
\end{figure*}

\subsection{Performance Sensitivity Analysis} 
\label{sec-performance}

In our paper, we have shown that KG data is very helpful to our model for both generation and coherence metrics. Here, we would examine how it affects the final performance. 
%Here, we investigate the influence of KG data amount and related model parameters. 

\subsubsection{Tuning the amount of KG data} The amount of available KG information directly affects the performance of various KG-enhanced methods. Here we examine how different methods perform with the varying amount of KG data. We select A-R2S+KG, KCGNN and PHVM as comparison methods.
We take 40\%, 60\%, 80\% and 100\% of the available KG data to generate four new KG training datasets, respectively. 
We utilize them together with the original review data to train our model and report the performance on the test set. The KG test set is fixed as original. As shown in Figure~\ref{fig-movie-varing-kg}, the performance of CETP gradually improves with the increasing of KG data, and CETP has achieved a consistent improvement over the other baselines with more than 40\% KG data. %It implies that CETP can also deal with sparsity case with fewer KG triples to some extent.

\subsubsection{Tuning the KG embedding size} For KG data, the embedding size is an important parameter to tune in real applications, which restricts the capacity of encoding KG information. Here, we vary the embedding size in the set \{64, 128, 256, 512\} and construct a similar evaluation experiment as that for the amount of KG data. As we can see from Figure~\ref{fig-movie-varing-es}, CETP is substantially better than the other baselines for all the four embedding sizes, which indicates the effectiveness of our model in extracting and encoding useful KG information. Another observation is that the performance of CETP gradually increases and the change is relatively stable. %	  More sensitivity results can be found in Appendix C.2.
%
%While, for another parameter, our experiment results show that stacking multi-attention blocks by six times give the best performance while the change with other numbers in \{1, 2, 4, 6\} is very small. We omit the results here. 

%Model sensitivity results on \textsc{Book} and \textsc{Electronic} datasets can be found in Appendix C.2.

\subsection{Qualitative Analysis}

Previous experiments have demonstrated the effectiveness of our model in generating semantically coherent review text. In this part, we qualitatively analyze why our model performs well. 

In Figure~\ref{fig-case}(a), we present the top 13 frequent subgraph schemas and their distributions from gold reviews and generated reviews of CETP and PHVM.
As we can see, the  distribution of subgraph schemas from CETP is closer to the real distribution (smaller MAE and RMSE results) than PHVM, indicating the effectiveness of our text planning mechanism. %The MAE and MSE for CETP and PHVM are \emph{0.0145 v.s. 0.0237} and \emph{0.0007 v.s. 0.0015}.
Furthermore,  Figure~\ref{fig-case}(b) and \ref{fig-case}(c) present two  movie reviews and corresponding text plans generated by CETP for a sample user. 
Note that, in Figure~\ref{fig-case}(b) and \ref{fig-case}(c), each generated sentence is verbalized from a generated subgraph (on colored background), and they are labelled with the same number. The number represent the order of the subgraph and sentence in their sequence generated by our model.

As we can see, for global coherence, CETP can capture inter-sentence entity distributions and generate similar aspect and content sketches compared with real reviews (\eg \emph{romance film (genre)}$ \rightarrow$ \emph{thomas mccarthy (director)}$ \rightarrow$ \emph{richard jenkins (actor)}), due to the effective text planning mechanism based on KG subgraphs.
For local coherence, the sentences are well verbalized through the intra-sentence correlation between entities in subgraphs (\eg \emph{thomas mccarthy} and \emph{richard kind}) and the connection of threading words (\eg \emph{stars} and \emph{performance}). 
Besides, CETP can capture  the preferred  relations and entities by the user about the two movies (\eg \emph{genre} and \emph{romance film}). This implies that the user-augmented KG data can provide important semantics for learning user preference.

% As we can see, our model is able to generate informative reviews by incorporating important entities from the proposed HKG. As a comparison, some copied entities (\eg \emph{tim burton}, \emph{joaquin phoenix}) occur in the form of the abbreviation (\eg \emph{mr. burton}, \emph{pheonix}) in real reviews. Moreover, our model has covered most of the aspects (with many overlapping aspect and opinion keywords) of the real reviews, which indicates the user's preference. Overall, with the constructed HKG, our model is able to produce informative and personalized reviews.

\section{Conclusion}

In this paper, we have presented a novel coherence-enhanced text
planning model for review generation. Our core idea is to utilize
KG subgraphs and their correlations to enhance local and global
coherence, respectively. KG subgraphs characterize the semantic
structure of intra-sentence entities which can naturally enforce the
local coherence since entities are tightly associated in subgraphs,
while subgraph sequence can capture complicated inter-sentence
correlations of entities to improve global coherence. The two kinds
of coherence have been modeled in a unified, principled text planning approach based on the HKG. Furthermore, we developed a
supervised copy mechanism to verbalize sentence based on KG subgraphs for further enhancing local coherence via copied threading
words. We have constructed extensive experiments on three real-world review datasets. The experimental results have demonstrated
the effectiveness of our model on review generation task in a series
of evaluation metrics. 

As future work, we will consider more kinds of external knowledge (e.g., WordNet) and investigate how our model could be applied to other domains.
\section*{Acknowledgement}
This work was partially supported by the National Natural Science Foundation of China under Grant No. 61872369 and 61832017, Beijing Academy of Artificial Intelligence (BAAI) under Grant No. BAAI2020ZJ0301, Beijing Outstanding Young Scientist Program under Grant No. BJJWZYJH012019100020098, the Fundamental Research Funds for the Central Universities, and the Research Funds of Renmin University of China under Grant No.18XNLG22 and 19XNQ047. Xin Zhao is the corresponding author.
\bibliographystyle{ACM-Reference-Format}
\bibliography{review_bib}


\begin{thebibliography}{48}

%%% the Web address.

\ifx \showCODEN    \undefined \def \showCODEN     #1{\unskip}     \fi
\ifx \showDOI      \undefined \def \showDOI       #1{#1}\fi
\ifx \showISBNx    \undefined \def \showISBNx     #1{\unskip}     \fi
\ifx \showISBNxiii \undefined \def \showISBNxiii  #1{\unskip}     \fi
\ifx \showISSN     \undefined \def \showISSN      #1{\unskip}     \fi
\ifx \showLCCN     \undefined \def \showLCCN      #1{\unskip}     \fi
\ifx \shownote     \undefined \def \shownote      #1{#1}          \fi
\ifx \showarticletitle \undefined \def \showarticletitle #1{#1}   \fi
\ifx \showURL      \undefined \def \showURL       {\relax}        \fi
% invisible to TeX
\providecommand\bibfield[2]{#2}
\providecommand\bibinfo[2]{#2}
\providecommand\natexlab[1]{#1}
\providecommand\showeprint[2][]{arXiv:#2}

\bibitem[\protect\citeauthoryear{Amazon}{Amazon}{2020}]%
        {imdb}
\bibfield{author}{\bibinfo{person}{Amazon}.} \bibinfo{year}{2020}\natexlab{}.
\newblock \bibinfo{title}{{IMD}b (Internet Movie Database)}.
\newblock \bibinfo{howpublished}{\url{https://www.imdb.com}}.
\newblock


\bibitem[\protect\citeauthoryear{Amplayo, Lim, and Hwang}{Amplayo
  et~al\mbox{.}}{2018}]%
        {AmplayoLH18}
\bibfield{author}{\bibinfo{person}{Reinald~Kim Amplayo},
  \bibinfo{person}{Seonjae Lim}, {and} \bibinfo{person}{Seung{-}won Hwang}.}
  \bibinfo{year}{2018}\natexlab{}.
\newblock \showarticletitle{Entity Commonsense Representation for Neural
  Abstractive Summarization}. In \bibinfo{booktitle}{\emph{NAACL}}.
  \bibinfo{pages}{697--707}.
\newblock


\bibitem[\protect\citeauthoryear{Bahdanau, Cho, and Bengio}{Bahdanau
  et~al\mbox{.}}{2015}]%
        {BahdanauCB14}
\bibfield{author}{\bibinfo{person}{Dzmitry Bahdanau},
  \bibinfo{person}{Kyunghyun Cho}, {and} \bibinfo{person}{Yoshua Bengio}.}
  \bibinfo{year}{2015}\natexlab{}.
\newblock \showarticletitle{Neural Machine Translation by Jointly Learning to
  Align and Translate}. In \bibinfo{booktitle}{\emph{ICLR}}.
\newblock


\bibitem[\protect\citeauthoryear{Bareket-Bojmel, Moran, and
  Shahar}{Bareket-Bojmel et~al\mbox{.}}{2016}]%
        {bareket2016strategic}
\bibfield{author}{\bibinfo{person}{Liad Bareket-Bojmel},
  \bibinfo{person}{Simone Moran}, {and} \bibinfo{person}{Golan Shahar}.}
  \bibinfo{year}{2016}\natexlab{}.
\newblock \showarticletitle{Strategic self-presentation on Facebook: Personal
  motives and audience response to online behavior}.
\newblock \bibinfo{journal}{\emph{Computers in Human Behavior}}
  \bibinfo{volume}{55} (\bibinfo{year}{2016}), \bibinfo{pages}{788--795}.
\newblock


\bibitem[\protect\citeauthoryear{Bartoli, De~Lorenzo, Medvet, and
  Tarlao}{Bartoli et~al\mbox{.}}{2016}]%
        {BartoliLMT16}
\bibfield{author}{\bibinfo{person}{Alberto Bartoli}, \bibinfo{person}{Andrea
  De~Lorenzo}, \bibinfo{person}{Eric Medvet}, {and} \bibinfo{person}{Fabiano
  Tarlao}.} \bibinfo{year}{2016}\natexlab{}.
\newblock \showarticletitle{Your paper has been accepted, rejected, or
  whatever: Automatic generation of scientific paper reviews}. In
  \bibinfo{booktitle}{\emph{International Conference on Availability,
  Reliability, and Security}}. Springer, \bibinfo{pages}{19--28}.
\newblock


\bibitem[\protect\citeauthoryear{Barzilay and Lapata}{Barzilay and
  Lapata}{2005}]%
        {BarzilayL05}
\bibfield{author}{\bibinfo{person}{Regina Barzilay} {and}
  \bibinfo{person}{Mirella Lapata}.} \bibinfo{year}{2005}\natexlab{}.
\newblock \showarticletitle{Modeling Local Coherence: An Entity-Based
  Approach}. In \bibinfo{booktitle}{\emph{ACL}}. \bibinfo{pages}{141--148}.
\newblock


\bibitem[\protect\citeauthoryear{Dalianis and Hovy}{Dalianis and Hovy}{1993}]%
        {DalianisH93}
\bibfield{author}{\bibinfo{person}{Hercules Dalianis} {and}
  \bibinfo{person}{Eduard~H. Hovy}.} \bibinfo{year}{1993}\natexlab{}.
\newblock \showarticletitle{Aggregation in Natural Language Generation}. In
  \bibinfo{booktitle}{\emph{EWNLG}} \emph{(\bibinfo{series}{Lecture Notes in
  Computer Science})}, Vol.~\bibinfo{volume}{1036}.
  \bibinfo{publisher}{Springer}, \bibinfo{pages}{88--105}.
\newblock


\bibitem[\protect\citeauthoryear{Devlin, Chang, Lee, and Toutanova}{Devlin
  et~al\mbox{.}}{2019}]%
        {DevlinCLT19}
\bibfield{author}{\bibinfo{person}{Jacob Devlin}, \bibinfo{person}{Ming{-}Wei
  Chang}, \bibinfo{person}{Kenton Lee}, {and} \bibinfo{person}{Kristina
  Toutanova}.} \bibinfo{year}{2019}\natexlab{}.
\newblock \showarticletitle{{BERT:} Pre-training of Deep Bidirectional
  Transformers for Language Understanding}. In
  \bibinfo{booktitle}{\emph{NAACL}}. \bibinfo{pages}{4171--4186}.
\newblock


\bibitem[\protect\citeauthoryear{Dong, Huang, Wei, Lapata, Zhou, and Xu}{Dong
  et~al\mbox{.}}{2017}]%
        {ZhouLWDHX17}
\bibfield{author}{\bibinfo{person}{Li Dong}, \bibinfo{person}{Shaohan Huang},
  \bibinfo{person}{Furu Wei}, \bibinfo{person}{Mirella Lapata},
  \bibinfo{person}{Ming Zhou}, {and} \bibinfo{person}{Ke Xu}.}
  \bibinfo{year}{2017}\natexlab{}.
\newblock \showarticletitle{Learning to Generate Product Reviews from
  Attributes}. In \bibinfo{booktitle}{\emph{EACL}}. \bibinfo{pages}{623--632}.
\newblock


\bibitem[\protect\citeauthoryear{Dubou{\'{e}} and McKeown}{Dubou{\'{e}} and
  McKeown}{2003}]%
        {DuboueM03}
\bibfield{author}{\bibinfo{person}{Pablo~Ariel Dubou{\'{e}}} {and}
  \bibinfo{person}{Kathleen~R. McKeown}.} \bibinfo{year}{2003}\natexlab{}.
\newblock \showarticletitle{Statistical Acquisition of Content Selection Rules
  for Natural Language Generation}. In \bibinfo{booktitle}{\emph{EMNLP}}.
\newblock


\bibitem[\protect\citeauthoryear{Elsner, Austerweil, and Charniak}{Elsner
  et~al\mbox{.}}{2007}]%
        {ElsnerAC07}
\bibfield{author}{\bibinfo{person}{Micha Elsner}, \bibinfo{person}{Joseph~L.
  Austerweil}, {and} \bibinfo{person}{Eugene Charniak}.}
  \bibinfo{year}{2007}\natexlab{}.
\newblock \showarticletitle{A Unified Local and Global Model for Discourse
  Coherence}. In \bibinfo{booktitle}{\emph{NAACL}}. \bibinfo{pages}{436--443}.
\newblock


\bibitem[\protect\citeauthoryear{Google}{Google}{2016}]%
        {freebase}
\bibfield{author}{\bibinfo{person}{Google}.} \bibinfo{year}{2016}\natexlab{}.
\newblock \bibinfo{title}{Freebase Data Dumps}.
\newblock
  \bibinfo{howpublished}{\url{https://developers.google.com/freebase/data}}.
\newblock


\bibitem[\protect\citeauthoryear{Grosz, Joshi, and Weinstein}{Grosz
  et~al\mbox{.}}{1995}]%
        {GroszJW95}
\bibfield{author}{\bibinfo{person}{Barbara~J. Grosz},
  \bibinfo{person}{Aravind~K. Joshi}, {and} \bibinfo{person}{Scott Weinstein}.}
  \bibinfo{year}{1995}\natexlab{}.
\newblock \showarticletitle{Centering: {A} Framework for Modeling the Local
  Coherence of Discourse}.
\newblock \bibinfo{journal}{\emph{Comput. Linguistics}} \bibinfo{volume}{21},
  \bibinfo{number}{2} (\bibinfo{year}{1995}), \bibinfo{pages}{203--225}.
\newblock


\bibitem[\protect\citeauthoryear{Guinaudeau and Strube}{Guinaudeau and
  Strube}{2013}]%
        {GuinaudeauS13}
\bibfield{author}{\bibinfo{person}{Camille Guinaudeau} {and}
  \bibinfo{person}{Michael Strube}.} \bibinfo{year}{2013}\natexlab{}.
\newblock \showarticletitle{Graph-based Local Coherence Modeling}. In
  \bibinfo{booktitle}{\emph{ACL}}. \bibinfo{pages}{93--103}.
\newblock


\bibitem[\protect\citeauthoryear{Guo, Lu, Cai, Zhang, Yu, and Wang}{Guo
  et~al\mbox{.}}{2018}]%
        {GuoLCZYW18}
\bibfield{author}{\bibinfo{person}{Jiaxian Guo}, \bibinfo{person}{Sidi Lu},
  \bibinfo{person}{Han Cai}, \bibinfo{person}{Weinan Zhang},
  \bibinfo{person}{Yong Yu}, {and} \bibinfo{person}{Jun Wang}.}
  \bibinfo{year}{2018}\natexlab{}.
\newblock \showarticletitle{Long Text Generation via Adversarial Training with
  Leaked Information}. In \bibinfo{booktitle}{\emph{AAAI}}.
  \bibinfo{pages}{5141--5148}.
\newblock


\bibitem[\protect\citeauthoryear{He and McAuley}{He and McAuley}{2016}]%
        {HeM16}
\bibfield{author}{\bibinfo{person}{Ruining He} {and} \bibinfo{person}{Julian
  McAuley}.} \bibinfo{year}{2016}\natexlab{}.
\newblock \showarticletitle{Ups and Downs: Modeling the Visual Evolution of
  Fashion Trends with One-Class Collaborative Filtering}. In
  \bibinfo{booktitle}{\emph{WWW}}. \bibinfo{pages}{507--517}.
\newblock


\bibitem[\protect\citeauthoryear{Hua and Wang}{Hua and Wang}{2019}]%
        {HuaW19}
\bibfield{author}{\bibinfo{person}{Xinyu Hua} {and} \bibinfo{person}{Lu Wang}.}
  \bibinfo{year}{2019}\natexlab{}.
\newblock \showarticletitle{Sentence-Level Content Planning and Style
  Specification for Neural Text Generation}. In
  \bibinfo{booktitle}{\emph{EMNLP-IJCNLP}}. \bibinfo{pages}{591--602}.
\newblock


\bibitem[\protect\citeauthoryear{Kingma and Ba}{Kingma and Ba}{2015}]%
        {KingmaB14}
\bibfield{author}{\bibinfo{person}{Diederik~P. Kingma} {and}
  \bibinfo{person}{Jimmy Ba}.} \bibinfo{year}{2015}\natexlab{}.
\newblock \showarticletitle{Adam: {A} Method for Stochastic Optimization}. In
  \bibinfo{booktitle}{\emph{ICLR}}.
\newblock


\bibitem[\protect\citeauthoryear{Lapata and Barzilay}{Lapata and
  Barzilay}{2005}]%
        {LapataB05}
\bibfield{author}{\bibinfo{person}{Mirella Lapata} {and}
  \bibinfo{person}{Regina Barzilay}.} \bibinfo{year}{2005}\natexlab{}.
\newblock \showarticletitle{Automatic Evaluation of Text Coherence: Models and
  Representations}. In \bibinfo{booktitle}{\emph{IJCAI}}.
  \bibinfo{pages}{1085--1090}.
\newblock


\bibitem[\protect\citeauthoryear{Li and Hovy}{Li and Hovy}{2014}]%
        {LiH14a}
\bibfield{author}{\bibinfo{person}{Jiwei Li} {and} \bibinfo{person}{Eduard~H.
  Hovy}.} \bibinfo{year}{2014}\natexlab{}.
\newblock \showarticletitle{A Model of Coherence Based on Distributed Sentence
  Representation}. In \bibinfo{booktitle}{\emph{EMNLP}},
  \bibfield{editor}{\bibinfo{person}{Alessandro Moschitti},
  \bibinfo{person}{Bo~Pang}, {and} \bibinfo{person}{Walter Daelemans}} (Eds.).
  \bibinfo{pages}{2039--2048}.
\newblock


\bibitem[\protect\citeauthoryear{Li, Li, Zhao, He, Wei, Yuan, and Wen}{Li
  et~al\mbox{.}}{2020a}]%
        {li2020knowledge}
\bibfield{author}{\bibinfo{person}{Junyi Li}, \bibinfo{person}{Siqing Li},
  \bibinfo{person}{Wayne~Xin Zhao}, \bibinfo{person}{Gaole He},
  \bibinfo{person}{Zhicheng Wei}, \bibinfo{person}{Nicholas~Jing Yuan}, {and}
  \bibinfo{person}{Ji{-}Rong Wen}.} \bibinfo{year}{2020}\natexlab{a}.
\newblock \showarticletitle{Knowledge-Enhanced Personalized Review Generation
  with Capsule Graph Neural Network}. In \bibinfo{booktitle}{\emph{CIKM}}.
  \bibinfo{publisher}{{ACM}}, \bibinfo{pages}{735--744}.
\newblock


\bibitem[\protect\citeauthoryear{Li, Zhao, Wen, and Song}{Li
  et~al\mbox{.}}{2019}]%
        {LiZWS19}
\bibfield{author}{\bibinfo{person}{Junyi Li}, \bibinfo{person}{Wayne~Xin Zhao},
  \bibinfo{person}{Ji{-}Rong Wen}, {and} \bibinfo{person}{Yang Song}.}
  \bibinfo{year}{2019}\natexlab{}.
\newblock \showarticletitle{Generating Long and Informative Reviews with
  Aspect-Aware Coarse-to-Fine Decoding}. In \bibinfo{booktitle}{\emph{ACL}}.
  \bibinfo{pages}{1969--1979}.
\newblock


\bibitem[\protect\citeauthoryear{Li, Zhang, and Chen}{Li
  et~al\mbox{.}}{2020b}]%
        {LiZC20}
\bibfield{author}{\bibinfo{person}{Lei Li}, \bibinfo{person}{Yongfeng Zhang},
  {and} \bibinfo{person}{Li Chen}.} \bibinfo{year}{2020}\natexlab{b}.
\newblock \showarticletitle{Generate Neural Template Explanations for
  Recommendation}. In \bibinfo{booktitle}{\emph{CIKM}},
  \bibfield{editor}{\bibinfo{person}{Mathieu d'Aquin}, \bibinfo{person}{Stefan
  Dietze}, \bibinfo{person}{Claudia Hauff}, \bibinfo{person}{Edward Curry},
  {and} \bibinfo{person}{Philippe Cudr{\'{e}}{-}Mauroux}} (Eds.).
  \bibinfo{publisher}{{ACM}}, \bibinfo{pages}{755--764}.
\newblock


\bibitem[\protect\citeauthoryear{Likert}{Likert}{1932}]%
        {likert1932technique}
\bibfield{author}{\bibinfo{person}{Rensis Likert}.}
  \bibinfo{year}{1932}\natexlab{}.
\newblock \showarticletitle{A technique for the measurement of attitudes.}
\newblock \bibinfo{journal}{\emph{Archives of psychology}}
  (\bibinfo{year}{1932}).
\newblock


\bibitem[\protect\citeauthoryear{Lin}{Lin}{2004}]%
        {Lin04}
\bibfield{author}{\bibinfo{person}{Chin-Yew Lin}.}
  \bibinfo{year}{2004}\natexlab{}.
\newblock \showarticletitle{ROUGE: A Package for Automatic Evaluation of
  Summaries}. In \bibinfo{booktitle}{\emph{Text Summarization Branches Out}}.
\newblock


\bibitem[\protect\citeauthoryear{Louis and Nenkova}{Louis and Nenkova}{2012}]%
        {LouisN12}
\bibfield{author}{\bibinfo{person}{Annie Louis} {and} \bibinfo{person}{Ani
  Nenkova}.} \bibinfo{year}{2012}\natexlab{}.
\newblock \showarticletitle{A Coherence Model Based on Syntactic Patterns}. In
  \bibinfo{booktitle}{\emph{EMNLP-CoNLL}}. \bibinfo{pages}{1157--1168}.
\newblock


\bibitem[\protect\citeauthoryear{Mani, Bloedorn, and Gates}{Mani
  et~al\mbox{.}}{1998}]%
        {mani1998using}
\bibfield{author}{\bibinfo{person}{Inderjeet Mani}, \bibinfo{person}{Eric
  Bloedorn}, {and} \bibinfo{person}{Barbara Gates}.}
  \bibinfo{year}{1998}\natexlab{}.
\newblock \showarticletitle{Using cohesion and coherence models for text
  summarization}. In \bibinfo{booktitle}{\emph{Intelligent Text Summarization
  Symposium}}. \bibinfo{pages}{69--76}.
\newblock


\bibitem[\protect\citeauthoryear{Mikolov, Sutskever, Chen, Corrado, and
  Dean}{Mikolov et~al\mbox{.}}{2013}]%
        {MikolovSCCD13}
\bibfield{author}{\bibinfo{person}{Tom{\'{a}}s Mikolov}, \bibinfo{person}{Ilya
  Sutskever}, \bibinfo{person}{Kai Chen}, \bibinfo{person}{Gregory~S. Corrado},
  {and} \bibinfo{person}{Jeffrey Dean}.} \bibinfo{year}{2013}\natexlab{}.
\newblock \showarticletitle{Distributed Representations of Words and Phrases
  and their Compositionality}. In \bibinfo{booktitle}{\emph{NIPS}}.
  \bibinfo{pages}{3111--3119}.
\newblock


\bibitem[\protect\citeauthoryear{Moryossef, Goldberg, and Dagan}{Moryossef
  et~al\mbox{.}}{2019}]%
        {MoryossefGD19}
\bibfield{author}{\bibinfo{person}{Amit Moryossef}, \bibinfo{person}{Yoav
  Goldberg}, {and} \bibinfo{person}{Ido Dagan}.}
  \bibinfo{year}{2019}\natexlab{}.
\newblock \showarticletitle{Step-by-Step: Separating Planning from Realization
  in Neural Data-to-Text Generation}. In \bibinfo{booktitle}{\emph{NAACL}},
  \bibfield{editor}{\bibinfo{person}{Jill Burstein}, \bibinfo{person}{Christy
  Doran}, {and} \bibinfo{person}{Thamar Solorio}} (Eds.).
  \bibinfo{pages}{2267--2277}.
\newblock


\bibitem[\protect\citeauthoryear{Ni, Li, and McAuley}{Ni et~al\mbox{.}}{2019}]%
        {NiLM19}
\bibfield{author}{\bibinfo{person}{Jianmo Ni}, \bibinfo{person}{Jiacheng Li},
  {and} \bibinfo{person}{Julian~J. McAuley}.} \bibinfo{year}{2019}\natexlab{}.
\newblock \showarticletitle{Justifying Recommendations using Distantly-Labeled
  Reviews and Fine-Grained Aspects}. In
  \bibinfo{booktitle}{\emph{EMNLP-IJCNLP}}. \bibinfo{pages}{188--197}.
\newblock


\bibitem[\protect\citeauthoryear{Ni, Lipton, Vikram, and McAuley}{Ni
  et~al\mbox{.}}{2017}]%
        {NiLVM17}
\bibfield{author}{\bibinfo{person}{Jianmo Ni}, \bibinfo{person}{Zachary~C.
  Lipton}, \bibinfo{person}{Sharad Vikram}, {and} \bibinfo{person}{Julian
  McAuley}.} \bibinfo{year}{2017}\natexlab{}.
\newblock \showarticletitle{Estimating Reactions and Recommending Products with
  Generative Models of Reviews}. In \bibinfo{booktitle}{\emph{IJCNLP}}.
  \bibinfo{pages}{783--791}.
\newblock


\bibitem[\protect\citeauthoryear{Ni and McAuley}{Ni and McAuley}{2018}]%
        {NiM18}
\bibfield{author}{\bibinfo{person}{Jianmo Ni} {and} \bibinfo{person}{Julian
  McAuley}.} \bibinfo{year}{2018}\natexlab{}.
\newblock \showarticletitle{Personalized Review Generation By Expanding Phrases
  and Attending on Aspect-Aware Representations}. In
  \bibinfo{booktitle}{\emph{ACL}}. \bibinfo{pages}{706--711}.
\newblock


\bibitem[\protect\citeauthoryear{Niu, Wu, Wang, et~al\mbox{.}}{Niu
  et~al\mbox{.}}{2019}]%
        {niu2019knowledge}
\bibfield{author}{\bibinfo{person}{Zheng-Yu Niu}, \bibinfo{person}{Hua Wu},
  \bibinfo{person}{Haifeng Wang}, {et~al\mbox{.}}}
  \bibinfo{year}{2019}\natexlab{}.
\newblock \showarticletitle{Knowledge Aware Conversation Generation with
  Explainable Reasoning over Augmented Graphs}. In
  \bibinfo{booktitle}{\emph{EMNLP-IJCNLP}}. \bibinfo{pages}{1782--1792}.
\newblock


\bibitem[\protect\citeauthoryear{Papineni, Roukos, Ward, and Zhu}{Papineni
  et~al\mbox{.}}{2002}]%
        {PapineniRWZ02}
\bibfield{author}{\bibinfo{person}{Kishore Papineni}, \bibinfo{person}{Salim
  Roukos}, \bibinfo{person}{Todd Ward}, {and} \bibinfo{person}{Wei{-}Jing
  Zhu}.} \bibinfo{year}{2002}\natexlab{}.
\newblock \showarticletitle{Bleu: a Method for Automatic Evaluation of Machine
  Translation}. In \bibinfo{booktitle}{\emph{ACL}}. \bibinfo{pages}{311--318}.
\newblock


\bibitem[\protect\citeauthoryear{Parveen, Mesgar, and Strube}{Parveen
  et~al\mbox{.}}{2016}]%
        {ParveenM016}
\bibfield{author}{\bibinfo{person}{Daraksha Parveen}, \bibinfo{person}{Mohsen
  Mesgar}, {and} \bibinfo{person}{Michael Strube}.}
  \bibinfo{year}{2016}\natexlab{}.
\newblock \showarticletitle{Generating Coherent Summaries of Scientific
  Articles Using Coherence Patterns}. In \bibinfo{booktitle}{\emph{EMNLP}}.
  \bibinfo{pages}{772--783}.
\newblock


\bibitem[\protect\citeauthoryear{Radford, Wu, Child, Luan, Amodei, and
  Sutskever}{Radford et~al\mbox{.}}{2019}]%
        {radford2019language}
\bibfield{author}{\bibinfo{person}{Alec Radford}, \bibinfo{person}{Jeffrey Wu},
  \bibinfo{person}{Rewon Child}, \bibinfo{person}{David Luan},
  \bibinfo{person}{Dario Amodei}, {and} \bibinfo{person}{Ilya Sutskever}.}
  \bibinfo{year}{2019}\natexlab{}.
\newblock \showarticletitle{Language models are unsupervised multitask
  learners}.
\newblock \bibinfo{journal}{\emph{OpenAI Blog}} \bibinfo{volume}{1},
  \bibinfo{number}{8} (\bibinfo{year}{2019}), \bibinfo{pages}{9}.
\newblock


\bibitem[\protect\citeauthoryear{Sha, Mou, Liu, Poupart, Li, Chang, and
  Sui}{Sha et~al\mbox{.}}{2018}]%
        {ShaMLPLCS18}
\bibfield{author}{\bibinfo{person}{Lei Sha}, \bibinfo{person}{Lili Mou},
  \bibinfo{person}{Tianyu Liu}, \bibinfo{person}{Pascal Poupart},
  \bibinfo{person}{Sujian Li}, \bibinfo{person}{Baobao Chang}, {and}
  \bibinfo{person}{Zhifang Sui}.} \bibinfo{year}{2018}\natexlab{}.
\newblock \showarticletitle{Order-Planning Neural Text Generation From
  Structured Data}. In \bibinfo{booktitle}{\emph{AAAI}}.
  \bibinfo{pages}{5414--5421}.
\newblock


\bibitem[\protect\citeauthoryear{Shao, Huang, Wen, Xu, and Zhu}{Shao
  et~al\mbox{.}}{2019}]%
        {ShaoHWXZ19}
\bibfield{author}{\bibinfo{person}{Zhihong Shao}, \bibinfo{person}{Minlie
  Huang}, \bibinfo{person}{Jiangtao Wen}, \bibinfo{person}{Wenfei Xu}, {and}
  \bibinfo{person}{Xiaoyan Zhu}.} \bibinfo{year}{2019}\natexlab{}.
\newblock \showarticletitle{Long and Diverse Text Generation with
  Planning-based Hierarchical Variational Model}. In
  \bibinfo{booktitle}{\emph{EMNLP-IJCNLP}}. \bibinfo{pages}{3255--3266}.
\newblock


\bibitem[\protect\citeauthoryear{Sutskever, Vinyals, and Le}{Sutskever
  et~al\mbox{.}}{2014}]%
        {SutskeverVL14}
\bibfield{author}{\bibinfo{person}{Ilya Sutskever}, \bibinfo{person}{Oriol
  Vinyals}, {and} \bibinfo{person}{Quoc~V. Le}.}
  \bibinfo{year}{2014}\natexlab{}.
\newblock \showarticletitle{Sequence to Sequence Learning with Neural
  Networks}. In \bibinfo{booktitle}{\emph{NIPS}}. \bibinfo{pages}{3104--3112}.
\newblock


\bibitem[\protect\citeauthoryear{Taheri, Gimpel, and Berger{-}Wolf}{Taheri
  et~al\mbox{.}}{2019}]%
        {TaheriGB19}
\bibfield{author}{\bibinfo{person}{Aynaz Taheri}, \bibinfo{person}{Kevin
  Gimpel}, {and} \bibinfo{person}{Tanya~Y. Berger{-}Wolf}.}
  \bibinfo{year}{2019}\natexlab{}.
\newblock \showarticletitle{Learning to Represent the Evolution of Dynamic
  Graphs with Recurrent Models}. In \bibinfo{booktitle}{\emph{WWW}}.
  \bibinfo{pages}{301--307}.
\newblock


\bibitem[\protect\citeauthoryear{Trivedi, Farajtabar, Biswal, and Zha}{Trivedi
  et~al\mbox{.}}{2019}]%
        {TrivediFBZ19}
\bibfield{author}{\bibinfo{person}{Rakshit Trivedi}, \bibinfo{person}{Mehrdad
  Farajtabar}, \bibinfo{person}{Prasenjeet Biswal}, {and}
  \bibinfo{person}{Hongyuan Zha}.} \bibinfo{year}{2019}\natexlab{}.
\newblock \showarticletitle{DyRep: Learning Representations over Dynamic
  Graphs}. In \bibinfo{booktitle}{\emph{ICLR}}.
\newblock


\bibitem[\protect\citeauthoryear{Vaswani, Shazeer, Parmar, Uszkoreit, Jones,
  Gomez, Kaiser, and Polosukhin}{Vaswani et~al\mbox{.}}{2017}]%
        {VaswaniSPUJGKP17}
\bibfield{author}{\bibinfo{person}{Ashish Vaswani}, \bibinfo{person}{Noam
  Shazeer}, \bibinfo{person}{Niki Parmar}, \bibinfo{person}{Jakob Uszkoreit},
  \bibinfo{person}{Llion Jones}, \bibinfo{person}{Aidan~N. Gomez},
  \bibinfo{person}{Lukasz Kaiser}, {and} \bibinfo{person}{Illia Polosukhin}.}
  \bibinfo{year}{2017}\natexlab{}.
\newblock \showarticletitle{Attention is All you Need}. In
  \bibinfo{booktitle}{\emph{NeurIPS}}. \bibinfo{pages}{5998--6008}.
\newblock


\bibitem[\protect\citeauthoryear{Wiseman, Shieber, and Rush}{Wiseman
  et~al\mbox{.}}{2017}]%
        {WisemanSR17}
\bibfield{author}{\bibinfo{person}{Sam Wiseman}, \bibinfo{person}{Stuart~M.
  Shieber}, {and} \bibinfo{person}{Alexander~M. Rush}.}
  \bibinfo{year}{2017}\natexlab{}.
\newblock \showarticletitle{Challenges in Data-to-Document Generation}. In
  \bibinfo{booktitle}{\emph{EMNLP}}. \bibinfo{pages}{2253--2263}.
\newblock


\bibitem[\protect\citeauthoryear{Yan and Han}{Yan and Han}{2002}]%
        {YanH02}
\bibfield{author}{\bibinfo{person}{Xifeng Yan} {and} \bibinfo{person}{Jiawei
  Han}.} \bibinfo{year}{2002}\natexlab{}.
\newblock \showarticletitle{gSpan: Graph-Based Substructure Pattern Mining}. In
  \bibinfo{booktitle}{\emph{ICDM}}. \bibinfo{publisher}{{IEEE} Computer
  Society}, \bibinfo{pages}{721--724}.
\newblock


\bibitem[\protect\citeauthoryear{Yang, Yih, He, Gao, and Deng}{Yang
  et~al\mbox{.}}{2015}]%
        {YangYHGD14a}
\bibfield{author}{\bibinfo{person}{Bishan Yang}, \bibinfo{person}{Wen{-}tau
  Yih}, \bibinfo{person}{Xiaodong He}, \bibinfo{person}{Jianfeng Gao}, {and}
  \bibinfo{person}{Li Deng}.} \bibinfo{year}{2015}\natexlab{}.
\newblock \showarticletitle{Embedding Entities and Relations for Learning and
  Inference in Knowledge Bases}. In \bibinfo{booktitle}{\emph{ICLR}}.
\newblock


\bibitem[\protect\citeauthoryear{Yu, Zhang, Wang, and Yu}{Yu
  et~al\mbox{.}}{2017}]%
        {YuZWY17}
\bibfield{author}{\bibinfo{person}{Lantao Yu}, \bibinfo{person}{Weinan Zhang},
  \bibinfo{person}{Jun Wang}, {and} \bibinfo{person}{Yong Yu}.}
  \bibinfo{year}{2017}\natexlab{}.
\newblock \showarticletitle{SeqGAN: Sequence Generative Adversarial Nets with
  Policy Gradient}. In \bibinfo{booktitle}{\emph{AAAI}}.
  \bibinfo{pages}{2852--2858}.
\newblock


\bibitem[\protect\citeauthoryear{Zang and Wan}{Zang and Wan}{2017}]%
        {ZangW17}
\bibfield{author}{\bibinfo{person}{Hongyu Zang} {and} \bibinfo{person}{Xiaojun
  Wan}.} \bibinfo{year}{2017}\natexlab{}.
\newblock \showarticletitle{Towards Automatic Generation of Product Reviews
  from Aspect-Sentiment Scores}. In \bibinfo{booktitle}{\emph{INLG}}.
  \bibinfo{pages}{168--177}.
\newblock


\bibitem[\protect\citeauthoryear{Zhao, He, Yang, Dou, Huang, Ouyang, and
  Wen}{Zhao et~al\mbox{.}}{2019}]%
        {zhao2019kb4rec}
\bibfield{author}{\bibinfo{person}{Wayne~Xin Zhao}, \bibinfo{person}{Gaole He},
  \bibinfo{person}{Kunlin Yang}, \bibinfo{person}{Hongjian Dou},
  \bibinfo{person}{Jin Huang}, \bibinfo{person}{Siqi Ouyang}, {and}
  \bibinfo{person}{Ji-Rong Wen}.} \bibinfo{year}{2019}\natexlab{}.
\newblock \showarticletitle{Kb4rec: A data set for linking knowledge bases with
  recommender systems}.
\newblock \bibinfo{journal}{\emph{Data Intelligence}} \bibinfo{volume}{1},
  \bibinfo{number}{2} (\bibinfo{year}{2019}), \bibinfo{pages}{121--136}.
\newblock


\end{thebibliography}

%\newpage
%\input{sec-appendix}

\end{document}